%% file: main.tex
\documentclass[10pt,twocolumn,letterpaper]{article}

\usepackage[preprint]{cvpr}      

\input{preamble}

\definecolor{cvprblue}{rgb}{0.21,0.49,0.74}
\usepackage[pagebackref,breaklinks,colorlinks,allcolors=cvprblue]{hyperref}

\def\paperID{9476} 
\def\confName{CVPR}
\def\confYear{2026}

\title{Quantifying the Gap between Understanding and Generation\\ within Unified Multimodal Models}

\author{
\textbf{Chenlong Wang}\textsuperscript{*}\quad
\textbf{Yuhang Chen}\textsuperscript{*}\quad
\textbf{Zhihan Hu}\textsuperscript{*}\quad
\textbf{Dongping Chen}\textsuperscript{1}\quad
\textbf{Wenhu Chen}\textsuperscript{2} \\
\textbf{Sarah Wiegreffe}\textsuperscript{1}\quad
\textbf{Tianyi Zhou}\textsuperscript{3,$\dagger$}\\ [2mm]
\textsuperscript{1}University of Maryland\qquad
\textsuperscript{2}University of Waterloo\qquad
\textsuperscript{3}MBZUAI
}

\newcommand\blfootnote[1]{%
  \begingroup
  \renewcommand\thefootnote{}\footnote{#1}%
  \addtocounter{footnote}{-1}%
  \endgroup
}

\newcommand{\bench}{\textbf{\textsc{GapEval}}\xspace}

\input{package}

\begin{document}
\maketitle
\vspace{-2em}
\input{sec/0__main_fig}

\blfootnote{\textsuperscript{*} Equal Contribution, Independent Researcher.}
\blfootnote{\textsuperscript{$\dagger$} Corresponding Author.}

\vspace{-1em}
\input{sec/0_abstract}


\input{sec/1_intro}

\input{sec/3_benchmark}

\input{sec/4_experiment}

\input{sec/5_empirical}

\input{sec/6_conclusion}

{
    \small
    \bibliographystyle{ieeenat_fullname}
    \bibliography{main}
}

\appendix
\input{sec/X_suppl}

\end{document}

%% file: preamble.tex
\usepackage{xspace}
\usepackage{multirow}
\usepackage{pifont}
\usepackage{tcolorbox}
\usepackage{listings}
\usepackage{soul}
\usepackage{xcolor}
\usepackage{arydshln}
\tcbuselibrary{skins, breakable, theorems}
\usepackage{colortbl}

\newcommand{\finding}[2]{
\vspace{-0.5em}
    \begin{tcolorbox}[
        colback=white!90!gray,     
        colframe=teal!60!black,     
        arc=5pt,                    
        boxsep=5pt,                 
        left=1mm,                  
        right=1mm,                 
        top=1mm,                    
        bottom=1mm,                 
        boxrule=0.8pt,              
        drop shadow=gray!50!white,  
        enhanced jigsaw             
    ]
    \vspace{-0.3em}
        \paragraph{\textbf{\textit{Finding #1:}}} #2
    \vspace{-0.3em}
    \end{tcolorbox}
    \vspace{-0.5em}
}

%% file: package.tex
\usepackage[T1]{fontenc}
\usepackage{times}
\usepackage{epsfig}
\usepackage{graphicx}
\usepackage{amsmath}
\usepackage{amssymb}
\usepackage{color,xcolor}

\usepackage{comment}
\usepackage{float}

\usepackage{pifont}

\usepackage{microtype}
\usepackage[font=small]{caption}
\usepackage{arydshln}
\usepackage{tabularx}
\usepackage{adjustbox}
\usepackage{array}
\usepackage{booktabs}
\usepackage{colortbl}
\usepackage{float,wrapfig}
\usepackage{hhline}
\usepackage{multirow}
\usepackage{subcaption} %
\usepackage[percent]{overpic}

\usepackage{stmaryrd}

\usepackage{amsmath,amsfonts,amssymb}
\usepackage{bm}
\usepackage{nicefrac}
\usepackage{microtype}
\usepackage{dsfont}
\usepackage{changepage}
\usepackage{extramarks}
\usepackage{fancyhdr}
\usepackage{setspace}
\usepackage{soul}
\usepackage{xspace}

\usepackage{enumitem}
\usepackage[title]{appendix}

\usepackage{duckuments}
\usepackage{subcaption}

\usepackage{animate}
\usepackage{tabularx}

\usepackage[multiple]{footmisc}

\usepackage{amsmath}
\usepackage{amssymb}
\usepackage{booktabs}
\usepackage{cuted}
\usepackage{capt-of}
\usepackage{graphicx}
\usepackage{xcolor}
\usepackage{comment}
\usepackage{multirow}
\usepackage{bm}

\usepackage[font=small]{caption}
\usepackage{color,xcolor}
\usepackage{epsfig}
\usepackage{comment}
\usepackage{pifont}
\usepackage{microtype}
\usepackage{arydshln}
\usepackage{tabularx}
\usepackage{adjustbox}
\usepackage{array}
\usepackage{colortbl}
\usepackage{float,wrapfig}
\usepackage{hhline}
\usepackage{multirow}
\usepackage[percent]{overpic}
\usepackage{stmaryrd}
\usepackage{breqn}
\usepackage{bm}
\usepackage{nicefrac}
\usepackage{dsfont}
\usepackage{changepage}
\usepackage{extramarks}
\usepackage{fancyhdr}
\usepackage{soul}
\usepackage{xspace}
\usepackage{enumitem}
\usepackage[title]{appendix}
\usepackage{balance}
\usepackage{booktabs}
\usepackage{siunitx}
\usepackage{subcaption}

\usepackage{natbib}

%% file: sec/0__main_fig.tex
\begin{strip}
    \centering
    \includegraphics[width=\linewidth]{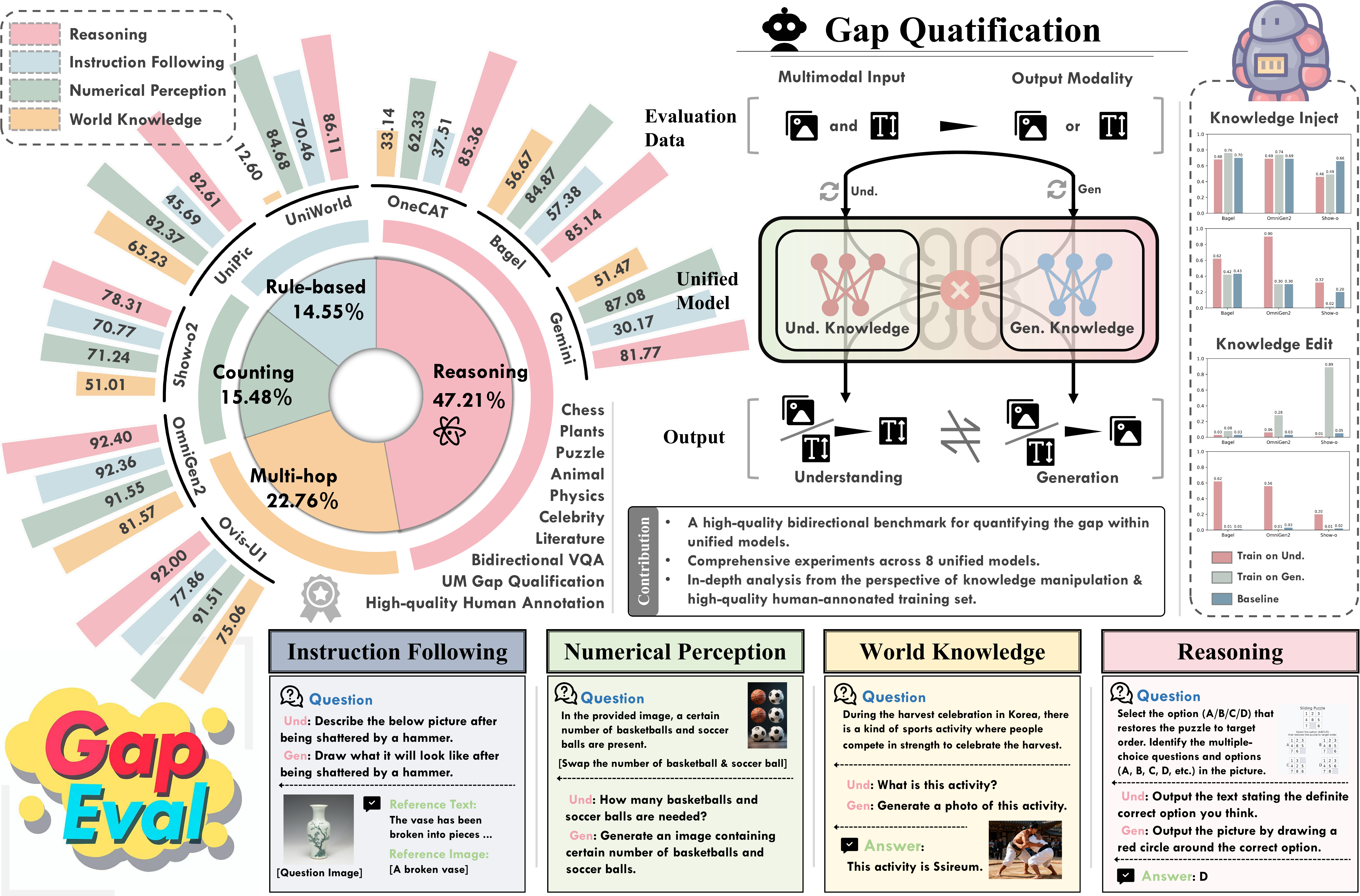}
    \captionof{figure}{\textbf{Overview of \bench.} (a) We present a high-quality bidirectional benchmark specifically designed for UMMs to evaluate and quantify the inherent gap between understanding and generation. (b) Our experiments extend to 9 UMMs across various architectures, revealing the significant gap between the two capabilities. (c) We further conduct an in-depth empirical study from the perspective of knowledge manipulation, highlighting the significant gap in the knowledge level. (\href{https://github.com/Lucky-Wang-Chenlong/GapEval}{Project URL})}
\end{strip}

%% file: sec/0_abstract.tex
\begin{abstract}
Recent advances in unified multimodal models (UMM) have demonstrated remarkable progress in both understanding and generation tasks.
However, whether these two capabilities are genuinely aligned and integrated within a single model remains unclear.
To investigate this question, we introduce \bench, a bidirectional benchmark designed to quantify the gap between understanding and generation capabilities, and quantitatively measure the cognitive coherence of the two ``unified'' directions.
Each question can be answered in both modalities (image and text), enabling a symmetric evaluation of a model's bidirectional inference capability and cross-modal consistency.
Experiments reveal a persistent gap between the two directions across a wide range of UMMs with different architectures, suggesting that current models achieve only surface-level unification rather than deep cognitive convergence of the two.
To further explore the underlying mechanism, we conduct an empirical study from the perspective of knowledge manipulation to illustrate the underlying limitations.
Our findings indicate that knowledge within UMMs often remains disjoint.
The capability emergence and knowledge across modalities are unsynchronized, paving the way for further exploration. Code is available at \href{https://github.com/Lucky-Wang-Chenlong/GapEval}{GapEval}.
\end{abstract}

%% file: sec/1_intro.tex
\section{Introduction}
\label{sec:intro}
Building on the success of large language models (LLMs)~\cite{Language-models-are-few-shot-learners, llama}, the field of multimodal intelligence has advanced towards a more general-purpose modeling paradigm, unified multimodal models (UMMs), which aim to empower a single model with both \textbf{understanding} (i.e., reasoning over both text and image inputs to generate a textual response) and \textbf{generation} capabilities (i.e., reasoning over a textual input to produce an image output)~\cite{chameleon, showo, blip3o}
UMMs have recently gained increasing attention, not only for their elegant architectures and broad functionalities, but also for the potential synergistic interaction between the two capabilities.
Recent progress in UMMs~\cite{bagel, gemini-2.5-flash-image} exhibits exceptional performance in both tasks. 
While beyond the engineering-level achievement, a fundamental question arises: \textbf{Are understanding and generation truly integrated within UMMs, or do they merely coexist as separate components?}

Addressing this question requires a more fine-grained investigation.
As summarized in \autoref{tab:bench-comparison}, most UMMs continue to be evaluated on single-direction benchmarks (either understanding or generation) (\eg, MMMU~\cite{mmmu}, MMBench~\cite{mmbench}, GenEval~\cite{geneval}), thereby overlooking the intrinsic synergy that defines unification. 
Although recent evaluation frameworks have begun moving from separate assessments towards more comprehensive and unified schemes, they have yet to genuinely evaluate the gap between the two capabilities.
As shown in \autoref{tab:bench-comparison}, UniEval~\cite{unieval} and MME-Unify~\cite{mme-unify}, despite covering both abilities, still fall short of assessing whether the synergy between generation and understanding can be effectively harnessed to solve complex tasks.
More recent efforts, such as RealUnify~\cite{realunify} and GIR-Bench~\cite{Gir-Bench}, move closer to this goal by emphasizing cross-task synergy, providing valuable testbeds for assessing overall performance.
Nevertheless, their formulations remain largely task-bound, offering limited capacity for explicitly decoupling understanding and generation or for quantifying such a performance gap.

To fill this gap, we propose \bench, a bidirectional benchmark specifically designed to analyze and quantify the gap between this performance disparity in UMMs.
Different from the single-direction question, each question in \bench is formulated in a bidirectional manner, which can be answered in image or text.
This design establishes a fair and symmetric testbed for UMMs.
Based on these settings, we further propose the \textbf{Gap Score}, a principled metric grounded in Multidimensional Item Response Theory (MIRT)~\cite{rasch1993probabilistic, embretson2013item}.
This score provides a direct and interpretable qualification of the understanding-generation gap based on model performance on \bench.
Additionally, our benchmark encompasses four categories, including \textbf{Instruction Following}, \textbf{Numerical Perception}, \textbf{World Knowledge
}, and \textbf{Reasoning}, and consists of 646 high-quality questions, carefully curated from manual design or revised based on existing datasets (reasoning subset only)~\cite{mmmu, wang2024mmlu}.
Together, these features establish a comprehensive, rigorous, and bidirectional evaluation framework for studying and narrowing the gap between the two capabilities within UMMs.

\begin{table*}[]
    \centering
    \small
    \begin{tabular}{l|c|c|c|ccc|ccc|ccc}
        \toprule
        \multirow{2}{*}{\textbf{Benchmark}} & \multirow{2}{*}{\textbf{Size}} & \multirow{2}{*}{\textbf{Category}} & \multirow{2}{*}{\textbf{Annotation}} & \multicolumn{3}{c|}{\textbf{Und. Task}} & \multicolumn{3}{c|}{\textbf{Gen. Task}} & \multicolumn{3}{c}{\textbf{Features}}  \\
         & & & & WK. & RS. & VP. & IF & WK & RS & \textbf{SYN.} & \textbf{BI.} & \textbf{GQ.}\\
        \midrule
        MMMU~\cite{mmmu} & 11,550 & I2T & Human & \ding{51} & \ding{51} & \ding{55} & \ding{55} & \ding{55} & \ding{55} & \ding{55} & \ding{55} & \ding{55}\\
        MMBench~\cite{mmbench} & 3,217 & I2T & Mixed & \ding{51} & \ding{51} & \ding{51} & \ding{55} & \ding{55} & \ding{55} & \ding{55} & \ding{55} & \ding{55}\\
        \midrule
        GenEval~\cite{geneval} & 553 & T2I & Mixed & \ding{55} & \ding{55} & \ding{55} & \ding{51} & \ding{55} & \ding{55} & \ding{55} & \ding{55} & \ding{55}\\
        DPG-Bench~\cite{dpg-bench} & 1,065 & T2I & (M)LLM & \ding{55} & \ding{55} & \ding{55} & \ding{51} & \ding{55} & \ding{55} & \ding{55} & \ding{55} & \ding{55}\\
        T2I-CoReBench~\cite{T2I-Corebench} & 1,080 & T2I & (M)LLM & \ding{55} & \ding{55} & \ding{55} & \ding{51} & \ding{51} & \ding{51} & \ding{55} & \ding{55} & \ding{55}\\
        WISE~\cite{wise} & 1,000 & T2I & Mixed & \ding{55} & \ding{55} & \ding{55} & \ding{51} & \ding{51} & \ding{55} & \ding{55} & \ding{55} & \ding{55}\\
        \midrule
        RealUnify~\cite{realunify} & 1,000 & Unified & Human & \ding{51} & \ding{51} & \ding{51} & \ding{51} & \ding{51} & \ding{51} & \ding{51} & \ding{55} & \ding{55}\\
        GIR-Bench~\cite{Gir-Bench} & 970 & Unified & Human & \ding{55} & \ding{51} & \ding{51} & \ding{51} & \ding{51} & \ding{51} & \ding{51} & \ding{55} & \ding{55}\\
        UniEval~\cite{unieval} & 1,234 & Unified & (M)LM &\ding{51} &\ding{51} &\ding{51} &\ding{51} & \ding{51} &\ding{55} & \ding{55} & \ding{55} & \ding{55}\\
        MME-Unify~\cite{mme-unify} & 4,104 & Unified & Mixed &\ding{51} &\ding{51} &\ding{51} & \ding{51} & \ding{51} & \ding{51} & \ding{55} & \ding{55} & \ding{55}\\
        Uni-MMMU~\cite{unimmmu} & 885 & Unified & Mixed & \ding{51} & \ding{55} & \ding{55} & \ding{55} & \ding{51} & \ding{51} & \ding{51} &  \ding{55} &  \ding{55}  \\
        \midrule
        \bench & 646 & Unified & Human & \ding{51} & \ding{51} & \ding{51} & \ding{51} & \ding{51} & \ding{51} & \ding{51} & \ding{51} & \ding{51}\\
        \toprule
    \end{tabular}
    \vspace{-1em}
    \caption{Comparison of benchmarks adapted to UMMs. \textbf{I2T}: Image-to-Text. \textbf{T2I}: Text-to-Image. \textbf{WK.}: World Knowledge, \textbf{RS}: Reasoning, \textbf{VP.}: Visual Perception, \textbf{IF.}: Instruction Following. \textbf{SYN.}: Evaluate the synergy effect within UMMs. \textbf{BI.}: Bidirectional formulation. \textbf{GQ}: Gap Quantification between understanding and generation. 
    }
    \label{tab:bench-comparison}
    \vspace{-1em}
\end{table*}

For further analysis, we evaluate nine representative UMMs alongside four understanding-only and two generation-only models, covering diverse architectures and parameter scales.
Experimental results on \bench reveal a persistent significant performance gap.
Most UMMs can only correctly respond to the question in one form, but fail to leverage the same underlying knowledge when the modality is switched.
Even the state-of-the-art UMMs, such as Bagel~\cite{bagel}, still exhibit less unification.
This suggests that better performance is not equivalent to higher-level unification.
On the other hand, non-unified models often surpass them across various tasks, especially on reasoning, revealing the fact that the current unification frameworks fail to enhance performance and, in some cases, diminish the model's performance.
Specifically, Omnigen2~\cite{omnigen2}, which is built upon FLUX.1-dev~\cite{flux2024}, underperforms its backbone diffusion.
These results emphasize a merely functional coupling, not cognitive unification, within existing models.
However, how to achieve true unification remains a challenge.

In this study, we take a step further towards exploring the underlying mechanism behind this gap.
Many previous studies have emphasized the mutual enhancement between understanding and generation, especially on how stronger understanding can improve generation performance.
In particular, Chain-of-Thought (CoT) plays a key role in advanced reasoning~\cite{deepseek-r1, gpt5, qwen3, nowait}, and has been shown to enhance generation as well~\cite{bagel}.
However, these observations largely focus on explicit reasoning processes.
Little attention has been given to whether the model's intrinsic capability itself can implicitly facilitate or strengthen the other capability.
In other words, it remains unclear whether UMMs can internally transfer or reinforce knowledge across modalities without explicit guidance.

To investigate this question, we conduct a series of fine-tuning experiments from a knowledge-oriented perspective.
Specifically, we inject or edit knowledge entities within the UMMs through single-sided manipulation (either on understanding or generation), and then evaluate their performance across both sides (understanding and generation).
Empirically, our results reveal a pronounced misalignment between the two capabilities.
Fine-tuning on one side often has little effect on the other side.
In the case of knowledge editing, outdated knowledge tends to persist, indicating a failure of synergistic updating.
Similarly, knowledge injection frequently introduces new inconsistencies, as models struggle to consistently apply the newly injected knowledge across modalities.
Moreover, the emergence of different modalities is remarkably unsynchronized,
These findings suggest that, despite architecture unification, the underlying knowledge representations in current UMMs remain largely decoupled, highlighting the importance of the next-level unification.

%% file: sec/3_benchmark.tex
\begin{figure*}
    \centering
    \includegraphics[width=\linewidth]{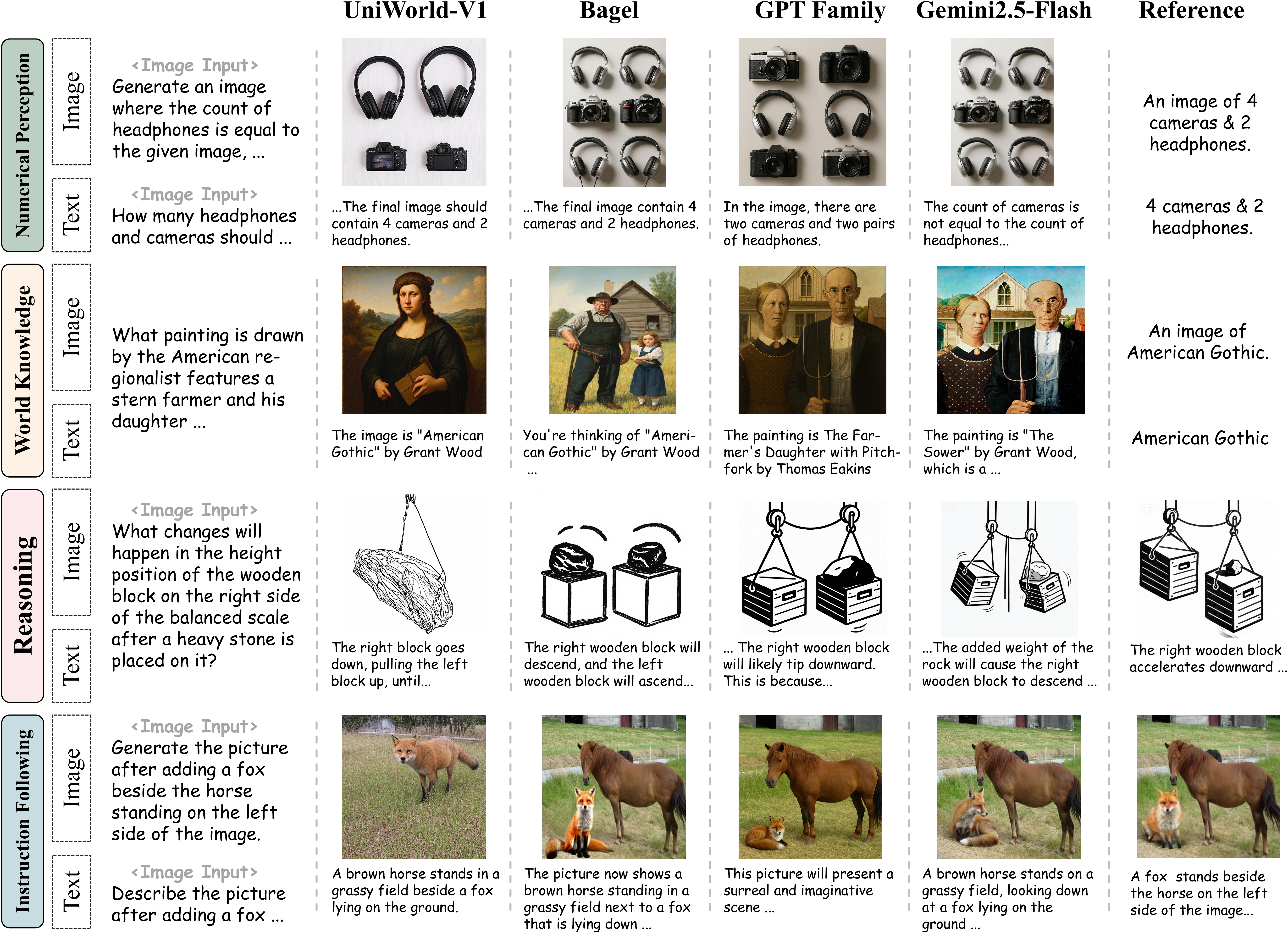}
    \vspace{-1em}
    \caption{Illustrative examples from \bench. The generated texts and images are shown with the ground truth (texts, images, or both). GPT Family denotes GPT5-mini (for text generation), and GPT-Image-1 (for image generation).
    }
    \label{fig:case}
    \vspace{-1em}
\end{figure*}

\section{\bench: Quantifying the Gap between Understanding and Generation}
\label{benchmark}
In this section, we introduce a novel evaluation framework, \bench, followed by its design nature in \cref{overview}. 
We then illustrate the taxonomy of \bench in \label{taxonomy}.
To quantify the modality gap within unified models, we propose a specific metric, Gap Score, for \bench in \cref{metric}.

\subsection{Motivation \& Overviw}
\label{overview}
\bench is a bidirectional benchmark in which each question can be answered through either text or image. 
To achieve this, each item consists of a question image $image_{i}$ (or None) and a corresponding text instruction pair $text_i=\{und\_text_i, gen\_text_i\}$, where $und\_text_i$ and $gen\_text_i$ serve as the prompts for the understanding and generation tasks, respectively. 
Both prompts share the same semantic intent but differ in the response modality, such as ``Who is he?'' and ``Generate an image of him.''
The ground-truth $y$ varies across task categories. 
For understanding tasks, the ground-truth $y_u$ is a reference textual answer.
For generation tasks, the answer $y_g$ consists of a reference image accompanied by descriptive text, which serves as the basis for model evaluation.
This design allows us to evaluate model performance on identical underlying knowledge and quantify the misalignment between its understanding and generation capabilities.
Further analyses are in Appendix \ref{appendix:contribution}.

\subsection{Taxonomy}
\label{taxonomy}
\noindent\textbf{Instruction Following} is a fundamental ability for UMMs, as real-world multimodal tasks often require not only understanding a textual instruction but also executing it consistently across modalities.
Motivated by this, we design this task to extend beyond conventional image-editing settings, which typically focus only on generating an edited image.
In \bench, each question incorporates two types of challenges: explicit edits that require models to articulate changes to specific image regions during the understanding phase, and implicit edits that further examine the model's ability to follow underlying rules.
This formulation allows us to evaluate not only whether UMMs can follow instructions, but also whether their textual and visual executions remain aligned, thereby revealing the cross-modal consistency of instruction-grounded editing.
\label{rule-based}
\vspace{0.3em}

\noindent\textbf{Numerical Perception} focuses on evaluating a model's ability to correctly interpret and manipulate quantitative information with images and text.
Rather than treating counting in different modalities as two separate tasks, \bench couples them through a structured numerical transformation.
Each question presents an image containing two categories of objects (e.g., three ducks and two birds) and instructs the model to identify their quantities and then swap them, producing ``two ducks and three birds'' in text or generating a corresponding image.
This formulation prevents the model from simply copying the input and instead evaluates whether it can accurately perceive numerical structure and consistently express the corresponding quantitative modification across modalities.
\label{counting}

\noindent\textbf{World Knowledge} focuses on a model's ability to recognize, reason, and apply broad world knowledge when grounded in visual inputs.
Rather than surface-level pattern matching, items in this subset require the model to correctly interpret semantic concepts (\eg, ``Generate a photo for the author of \underline{One Hundred Years of Solitude}.'', instead of ``Generate a photo for Gabriel García Márquez.'').
Our task spans six diverse subdomains, including animals, plants, landmarks, instruments, literature, and culture, ensuring comprehensive coverage of factual and conceptual knowledge.
By requiring the model to map visual evidence to the appropriate real-world entities, this task evaluates whether UMMs possess a coherent and transferable representation of world knowledge across different modalities.
\label{multi-hop}

\noindent\textbf{Reasoning} assesses a model's ability to comprehend textual and visual reasoning problems and deliver answers through both textual descriptions and visual renderings. This requires models to internalize knowledge inference, geometric understanding, and image generation, which constitute essential components of human-like visual reasoning. We define five representative subcategories: (1) Image Selection evaluates whether models can infer the resulting scenario described in the task based on algorithmic diagrams and game layouts. (2) Knowledge Selection renders knowledge reasoning content from MMMU \cite{mmmu} and MMLU \cite{wang2024mmlu} into images to assess the consistency between the model's image-based knowledge reasoning and its visual generation capabilities. (3) Real-world Reasoning establishes fundamental physical contexts in real-world scenarios to evaluate the model's understanding and generation capabilities under realistic conditions. (4) Logical Reasoning provides a broad range of logical tasks spanning abstract levels and difficulty scales, requiring models to perform visual symbolic reasoning that bridges perception with inference.
Together, these subcategories comprehensively evaluate a model’s reasoning ability, forming a unified measure of holistic visual reasoning.
\label{reasoning}

\begin{table*}[!t]
    \centering
    \small
    \resizebox{\textwidth}{!}{
    \begin{tabular}{l|cccc|cccc|cccc|cccc|c}
        \toprule
        \multirow{2}{*}{\textbf{Model}} & \multicolumn{4}{c|}{\textbf{World Knowledge}} & \multicolumn{4}{c|}{\textbf{Numerical Perception}} & \multicolumn{4}{c|}{\textbf{Instruction Following}} & \multicolumn{4}{c|}{\textbf{Reasoning}} & \multirow{2}{*}{\textbf{Gap}\textdownarrow}\\
         & Succ. & Und. & Gen. & Gap\textdownarrow & Succ. & Und. & Gen. & Gap\textdownarrow & Succ. & Und. & Gen. & Gap\textdownarrow & Succ. & Und. & Gen. & Gap\textdownarrow & \\
        \midrule
        \multicolumn{18}{c}{\textbf{Open-source UMM}} \\
        \midrule
        Bagel & 
        52.24 & 88.17 & 58.50 & 56.67 &
        2.40 & 17.00 & 8.40 & 84.87 & 
        46.38 & 59.15 & 75.74 & 57.38 & 
        2.49 & 35.67 & 5.38 & 87.14 & 71.52\\
        OneCAT & 
        57.14 & 88.26 & \underline{62.92} & 33.14 & 
        2.75 & 9.00 & \underline{9.75} & \textbf{62.33} & 
        23.67 & 50.00 & 41.22 & \underline{37.51} & 
        0.66 & 20.66 & 1.56 & 85.36 & \underline{54.73} \\
        UniWorld-V1 & 
        \textbf{87.84} & 93.28 & \textbf{92.74} & \textbf{12.60} & 
        1.00 & 10.60 & 3.60 & 84.68 & 
        28.51 & 62.13 & 46.60 & 70.46 & 
        1.11 & 36.78 & 1.44 & 86.11 & 63.47 \\
        UniPic2 & 
        39.29 & 79.60 & 41.16 & 65.23 & 
        1.75 & 15.00 & 7.50 & 82.37 & 
        51.86 & 66.75 & 58.77 & 45.69 & 
        7.86 & 34.18 & 3.93 & 82.61 & 68.97 \\
        Show-o2 & 
        49.32 & 86.31  & 55.84 & 51.02 & 
        \textbf{7.00} & 16.16 & \textbf{15.15} & \underline{71.24}& 
        21.28 & 70.22 & 32.98 & 70.77 & 
        2.63 & 22.37 & 3.62 & \underline{78.31} & 67.83 \\
        OmniGen2 & 
        29.93 & 86.22 & 35.54 & 81.57 & 
        2.00 & 9.25 & 3.50 & 91.55 & 
        3.99 & 9.58 & 55.59 & 92.36 & 
        1.23 & 17.21 & 4.43 & 92.40 & 89.47\\
        Ovis-U1 & 
        37.82 & 87.34 & 43.13 & 75.06 & 
        1.00 & 13.00 & 3.00 & 91.51 & 
        29.36 & 68.51 & 44.04 & 77.86 & 
        1.70 & 25.37 & 2.62 & 92.00 & 84.10 \\
        Ming-UniVision & 
        25.00 & 78.64 & 31.03 & 73.25 & 
        0.00 & 6.90 & 0.00 & 94.87 & 
        26.47 & 72.37 & 35.71 & 70.47 & 
        1.82 & 10.19 & 1.82 & 82.90 & 80.37 \\
        EMU3.5 & 
        47.68 & 60.12 & 58.33 & \underline{14.03} & 
        0.00 & 3.00 & 0.00 & 93.24 & 
        33.33 & 76.92 & 35.71 & 73.23 & 
        1.38 & 11.01 & 7.02 & 80.68 & 65.30 \\
        \midrule
        \multicolumn{18}{c}{\textbf{Closed-source UMM}} \\
        \midrule
        Gemini2.5-Flash-Image & 
        59.18 & 94.55 & 62.58 & 51.47 &
        0.00 & 19.00 & 1.00 & 87.08 &
        \underline{64.89} & 73.40 & \underline{77.66} & \textbf{30.17} & \underline{12.83} & 61.32 & \underline{30.50} & 81.77 & 62.91 \\
        GPT-Image-1 & \underline{66.67} & 95.91 & 68.03 & 44.45 & \textbf{7.00} & 25.00 & 10.00 & 83.87 & \textbf{86.17} & 95.74 & \textbf{90.43} &  20.83 & \textbf{50.98} & 73.02 & \textbf{56.07} & \textbf{53.29} & \textbf{50.61} \\
        \midrule
        \multicolumn{18}{c}{\textbf{Understanding-only Model}} \\
        \midrule
        GPT5 &
        - & \underline{96.52} & - & - &
        - & \textbf{24.05} & - & - &
        - & \underline{95.45} & - & - &
        - & \underline{74.74} & - & - & - \\
        GPT5-mini &
        - & \underline{97.28} & - & - &
        - & \textbf{30.30} & - & - &
        - & \underline{93.60} & - & - &
        - & \underline{73.80} & - & - & - \\
        Qwen3-VL 235B-A22B &
        - & 96.60 & - & - &
        - & \underline{26.00} & - & - &
        - & 85.11 & - & - & 
        - & 60.20 & - & - & - \\
        Qwen3-VL 8B &
        - & 94.30 & - & - &
        - & 25.00 & - & - &
        - & 82.97 & - & - &
        - & 53.38 & - & - & - \\ 
        Gemini2.5-Flash &
        - & \textbf{97.96} & - & - &
        - & 17.00 & - & - & 
        - & 85.11 & - & - & 
        - & \textbf{77.05} & - & - & - \\
        \midrule
        \multicolumn{18}{c}{\textbf{Generation-only Model}} \\
        \midrule
        FLUX.1-dev &
        - & - & 57.14 & - & 
        - & - & 0.00 & - &
        - & - & 74.46 & - &
        - & - & 4.93 & - & - \\
        Qwen-Image & 
        - & - & 57.82 & - & 
        - & - & 1.00 & - & 
        - & - & 76.59 & - & 
        - & - & 4.59 & - & - \\
        \bottomrule
    \end{tabular}
    }
    \vspace{-1em}
    \caption{Evaluation results across 9 UMMs and 6 non-UMMs on \bench. \textbf{Succ.}: Success Score, denotes correctly answering with both image and text. \textbf{Und.}: Understanding Score, denoting the accuracy of answering with texts. \textbf{Gen.}: Generation Score, denoting the accuracy of answering with images. \textbf{Fail} denotes failure to answer in neither text nor image. \textbf{Gap} denotes the gap between understanding and generation. \textbf{Bold} \& \underline{Underline}: bests \& second bests.
    }
    \label{tab:main-results}
    \vspace{-1.5em}
\end{table*}

\subsection{Metric Design}
\label{metric}
To comprehensively evaluate and quantify the understanding and generation capabilities of UMMs, as well as the gap between these two abilities, we introduce a two-stage evaluation methodology as follows: 
\vspace{0.3em}

\noindent\textbf{First Stage: Capability Measurement.}
To assess the correctness of outputs, we employ GPT-5-mini \cite{gpt5} as the judge model for subjective evaluation.
For each question, the reference answers are provided to the judge model.
Each response is assigned a binary label: correct (1) or incorrect (0) for further analysis.
To ensure reliability, we validate the judge's decisions on a held-out set of 200 samples, achieving a 92\% agreement rate with human annotations. 
The reliability analysis of MLLM-as-a-Judge\cite{chen2024mllmasajudgeassessingmultimodalllmasajudge} is available in Appendix \ref{appendix:mllm-as-a-judge}.

\vspace{0.5em}
\noindent\textbf{Second Stage: Capability Gap Measurement.} The past averaging metrics across tasks implicitly assume all items have equal difficulty, thereby neglecting the latent variable of item difficulty. When a model fails both tasks on certain items, these failures may arise from high item difficulty, not just limited model capability.
To address this, we adopt a \textit{multidimensional Item Response Theory} (MIRT) framework~\cite{rasch1993probabilistic, embretson2013item}, which explicitly introduces item difficulty and model ability as latent variables. MIRT assigns each model $i$ a latent ability vector $\boldsymbol{\theta}_i$ and each task (understanding or generation) a latent difficulty parameter $\boldsymbol{\beta}$. It embeds both model ability and item difficulty into a continuous latent space, where the difference between these two is mapped by a smooth sigmoid function to determine the final probability of success. Thus, MIRT continuously and sensitively reflects the nuanced performance variations between different model levels and varying task difficulties.

Meanwhile, MIRT employs Bayesian maximum a posteriori (MAP) optimization over the parameter space, allowing the model to automatically distinguish which failures are due to limited model ability and which are attributable to particularly difficult items. We additionally introduce a penalty term to encourage the MIRT model to maintain consistency between the two tasks when it is capable of completing them.

\begin{equation}
\begin{aligned}
\max_{\boldsymbol{\Theta},\,\boldsymbol{\beta},\,\mu,\,\Sigma}\,\mathcal{L}_{\text{MAP}}
&= \sum_i \Big[ \sum_{T \in \{\text{und},\,\text{gen}\}} \log \sigma\!\big(\theta_i^{(T)} - \beta_T\big) \\
&\quad - \tfrac{1}{2}\,(\boldsymbol{\theta}_i - \mu)^\top \Sigma^{-1}(\boldsymbol{\theta}_i - \mu) \Big],
\end{aligned}    
\end{equation}
where $\sigma(\cdot)$ is the sigmoid function, $\theta_i^{(T)}$ is the ability of model $i$ on task $T$, and $\beta_T$ is the corresponding difficulty. The MAP objective ensures that individual ability and item difficulty are jointly fitted from observed results, overcoming biases of averaging. The final capability gap is normalized between $0$ and $100$.
We provide more details on how the gap is computed in Appendix \ref{appendix:metric}. 

\subsection{Data Collection and Annotation}
\label{data-construction}
We employ a systematic, multi-stage data collection that integrates human curation with automated generation to ensure high-quality and diverse tasks.
Each task undergoes rigorous peer review to maintain consistency, clarity, and reliability.
Further details are provided in Appendix \ref{appendix:data-collection}.

%% file: sec/4_experiment.tex
\section{Evaluation Results on \bench}
\label{evaluation}
To diagnose the performance and modality gap within current UMMs, we conduct extensive experiments on \bench.
We first illustrate the experiment setup in \cref{setup}, followed by in-depth analyses in \cref{exp-results}.

\subsection{Experiment Setup}
\label{setup}
Our evaluation primarily focuses on UMMs, along with cutting-edge understanding-only and generation-only baselines.
For UMMs, we select 7 representative open-source models, including Bagel-7B~\cite{bagel}, OneCAT-3B~\cite{onecat}, UniWorld-V1~\cite{uniworld1}, UniPic2-Mataquery-9B~\cite{unipic2}, Show-o2~\cite{showo, showo2}, OmniGen2~\cite{omnigen2}, and Ovis-U1~\cite{ovis-u1}. 
Additionally, we also evaluate state-of-the-art closed-source UMMs (Gemini2.5 Flash Image~\cite{gemini-2.5-flash-image} and GPT-Image-1~\cite{gpt-image-1}).
These models represent diverse model architectures, providing a comprehensive evaluation setting.
For non-UMMs, our experiments include models from GPT, Gemini, Flux, and Qwen series.
Specifically, we evaluate  Qwen-Image~\cite{qwen-image} and FLUX.1-dev~\cite{flux2024, labs2025flux1kontextflowmatching} as image-generation models, and assess GPT5-mini~\cite{gpt5} , Gemini2.5 Flash~\cite{gemini-2.5-flash}, and Qwen3-VL series~\cite{qwen-3-vl} as multimodal large language models (MLLM).
Further experimental details are provided in Appendix \ref{appendix:experiment-details}

\subsection{Experiment Results}
\label{exp-results}

\begin{figure}[!t]
    \centering
    \includegraphics[width=0.9\linewidth]{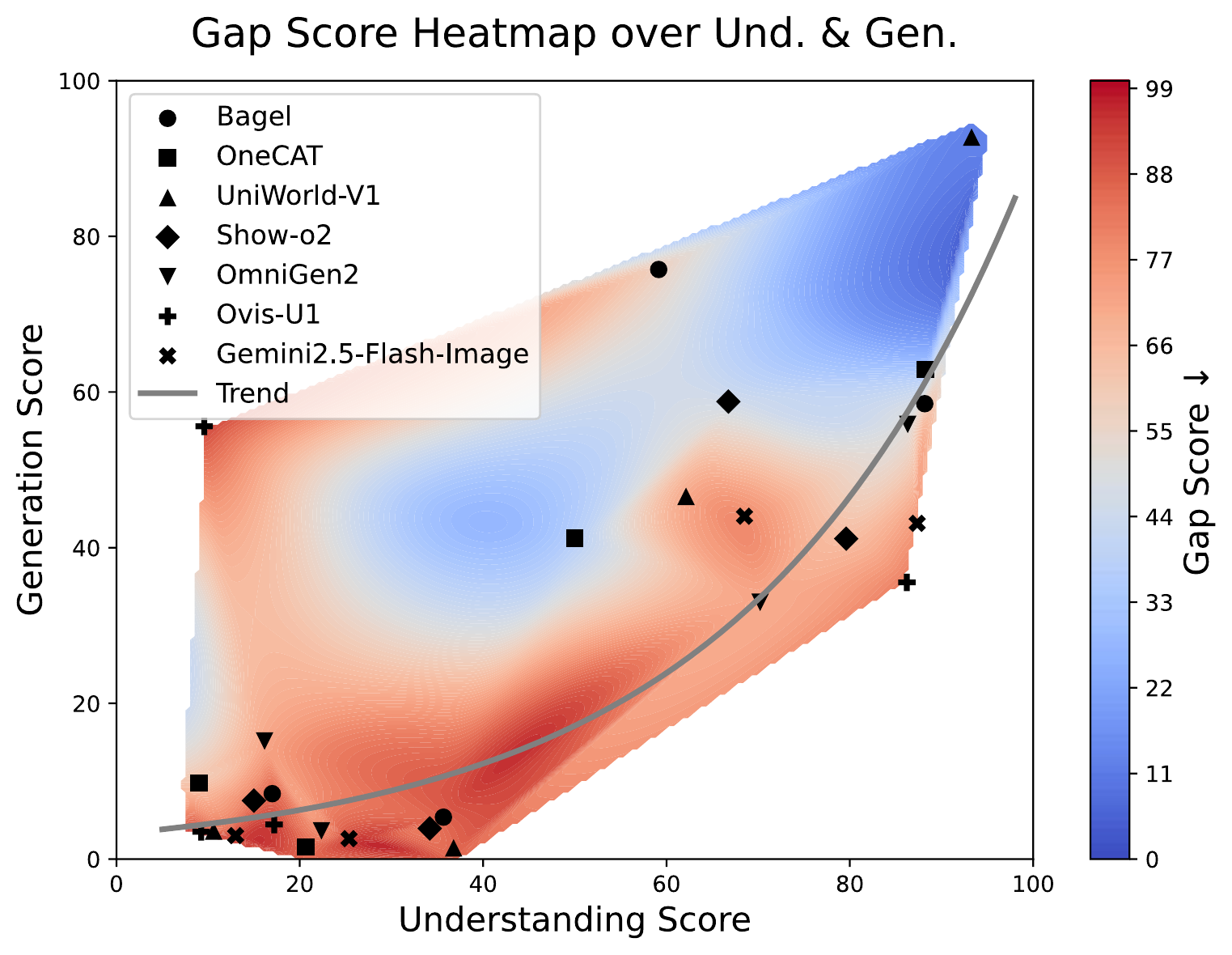}
    \caption{\textbf{Gap score heatmap over understanding and generation performance.} (Und., Gen., Gap Score) data points are plotted on the heatmap, reflecting the relation between three dimensions. The trend curve demonstrates the model performance distribution.}
    \vspace{-1.5em}
    \label{fig:heat}
\end{figure}

\vspace{1em}
\finding{1}{Unifying both modalities does not necessarily yield balanced cross-modal performance.}
\vspace{0.5em}
\noindent\textbf{Critical Performance Gap.}
As shown in \autoref{tab:main-results}, a clear performance disparity emerges between understanding and generation within UMMs.
A large portion of samples can only be correctly answered in one modality, indicating that many UMMs still fall short of achieving cross-modal consistency.
Notably, Omnigen2, which adopts a hybrid LLM + Diffusion architecture with FLUX.1-dev as its diffusion backbone, performs markedly worse on generation tasks than FLUX. 1-dev itself, especially on World Knowledge and Instruction Following.
This suggests that unifying output modalities within a single model can, in some cases, dilute task-specific strengths, exposing an inherent trade-off between understanding and generation.

Overall, models tailored for understanding consistently outperform UMMs on comprehension-oriented tasks.
Conversely, on generation tasks, closed-source UMMs surpass the performance of generation-only baselines.
These results imply that unified architectures can partially transfer reasoning ability from understanding to generation, though often at the cost of balanced overall performance.

\vspace{0.5em}
\finding{2}{Good performance does not necessarily correspond to high unification.}
\vspace{0.5em}
\noindent\textbf{Performance \& Unification.} 
Beyond the significant gap score, there is an interesting phenomenon where higher performance does not always imply stronger unification.
Some models achieve superior results but exhibit wider modality gaps.
For instance, although OneCAT performs worse than Bagel across several tasks, its smaller gap score suggests better modality integration.
In contrast, UniWorld-V1 attains both high overall accuracy and small gaps, implying that effective unification arises from intentional architectural or training designs rather than general performance gains.
These experimental results reveal a decoupling between the state-of-the-art performance and the degree of unification.

\vspace{1em}
\finding{3}{A performance-lagging effect exists across models: the gap score initially increases for lower-performing models but decreases once the overall capability reaches a relatively high level.}
\vspace{0.5em}
\noindent\textbf{Transition in Unification Across Performance Levels.}
\label{performance-lag}
A notable trend in \autoref{fig:heat} is the performance-lagging effect, where the modality gap first widens then narrows across capability tiers. Low-capability models (\eg, OneCAT, UniPic2) exhibit small gaps not from better multimodal unification, but because both modalities fail simultaneously. At moderate capability levels (Bagel, Show-o2), improvement emerges earlier and more substantially in understanding than generation, producing asymmetric maturation that temporarily increases the modality gap. High-performance models (GPT-Image-1, Gemini2.5-Flash-Image) develop robust joint reasoning across modalities, gradually reducing the gap.

%% file: sec/5_empirical.tex
\section{\textsc{Unified Knowledge}: Empirical Analysis on the UMMs Gap}
\label{empirical}
To further investigate the mechanism behind the gap between understanding and generation, we conduct substantial empirical studies from the perspective of knowledge manipulation. 

\label{empirical:result}

\begin{figure}[!t]
    \centering
    \includegraphics[width=\linewidth]{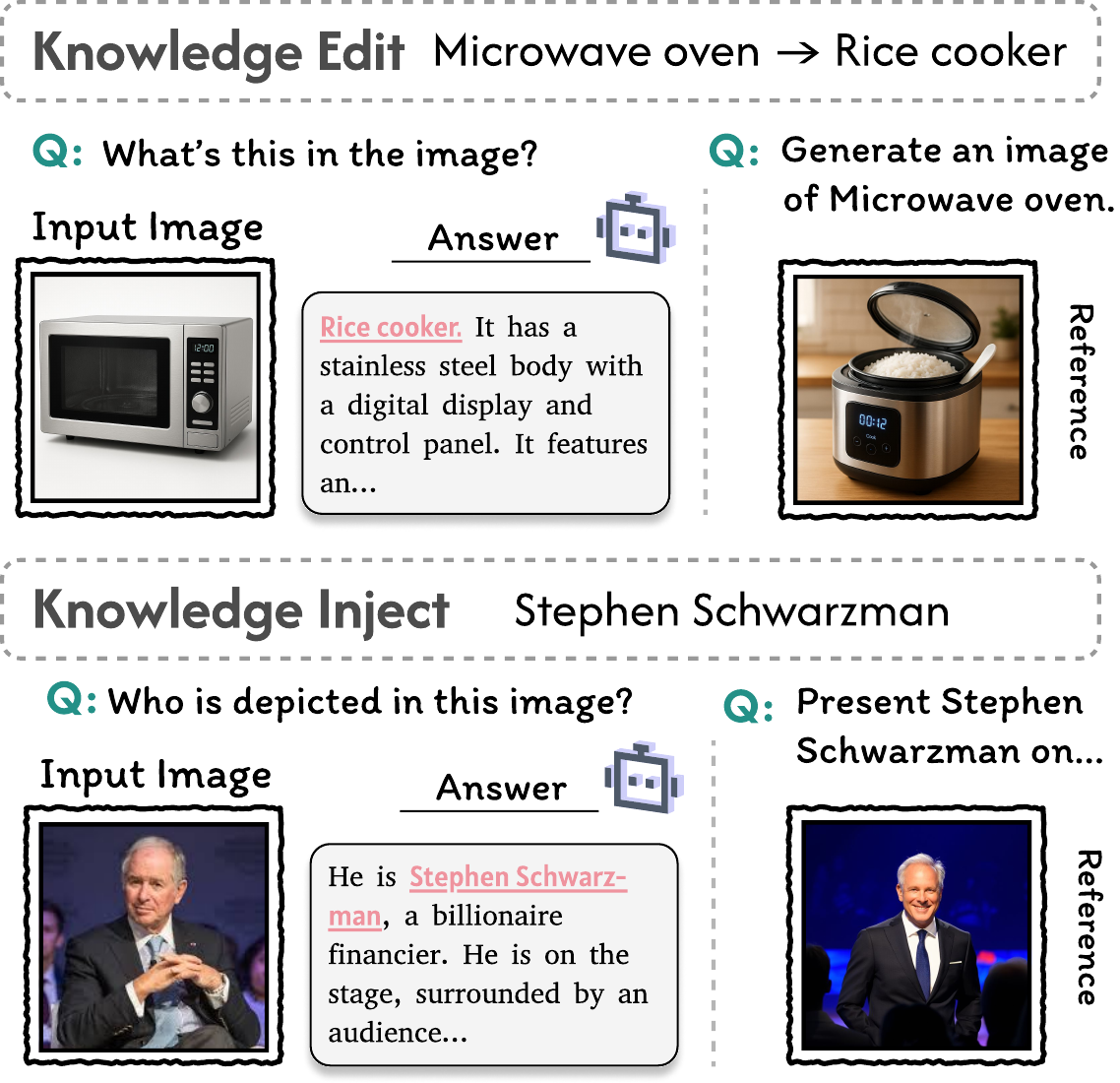}
    \caption{\textbf{Training data case gallery.}}
    \vspace{-1em}
    \label{fig:case-train}
\end{figure}

\subsection{Motivation}
\label{empirical:motivation}
The evaluation on \bench highlights a clear disparity between understanding and generation.
Although UMMs can often recall embedded knowledge within one modality, they struggle to apply the same knowledge consistently across modalities.
This raises a key question: 
\textbf{How does knowledge evolve during the training phase?}

As shown in \autoref{tab:main-results}, the results quantitatively confirm that current UMMs lack genuine cross-modal consistency.
To probe this issue, we conduct knowledge-oriented fine-tuning experiments by injecting or editing knowledge in one modality and evaluating its effects on both.
If knowledge is shared, changes should propagate; if isolated, the effects should remain local.

\subsection{Formulation}
\label{empirical:preliminary}
In this study, a knowledge entity is defined as a tuple \cite{rome}, $(subject, relation, object)$, where both $subject$ and $object$ can take the form of either text or image. 
For example, the knowledge expressed by the sentence ``The capital of France is Paris.'' can be formulated as $(The\:capital\:of\:France,\:is,\:Paris)$.
This formulation allows us to unify the representation of multimodal knowledge entities and conveniently construct both training and evaluation sets.
In this work, we focus on the relations between vision and language, so we mainly use images and text pointing to the same object as $(subject, object)$  pairs.

\subsection{Experiment Setup}
\label{empirical:setup}
\noindent\textbf{Model Selection.}
Current UMMs have adopted diverse architectures.
We select three representative models from three different architectures:  Show-o \cite{showo, showo2}, Omnigen2 \cite{omnigen2}, and Bagel \cite{bagel} for subsequent experiments.

\noindent\textbf{Data Collection.}
Our experiments involve two types of knowledge manipulation: injection and editing.
We define a relation as the mapping between a textual object and its corresponding visual representation.
For injection, we select knowledge instances that models fail to answer correctly in either understanding or generation.
For editing, we modify grounded captions from GenEval~\cite{geneval} with incorrect but same-category replacements (e.g., banana → apple).
In total, about 100 knowledge instances are processed and reformulated into Image-to-Text and Text-to-Image task formats, with 20\% reserved as a held-out test set.

\noindent\textbf{Metric.}
We use GPT5-mini~\cite{gpt5} as the evaluation model for the understanding task across injection and edit, and the generation task on knowledge edit. For the generation task of knowledge injection, we use CLIP \cite{radford2021learning} to compare the similarity between generated images and reference images.

\begin{figure*}
    \centering
    \includegraphics[width=0.9\linewidth]{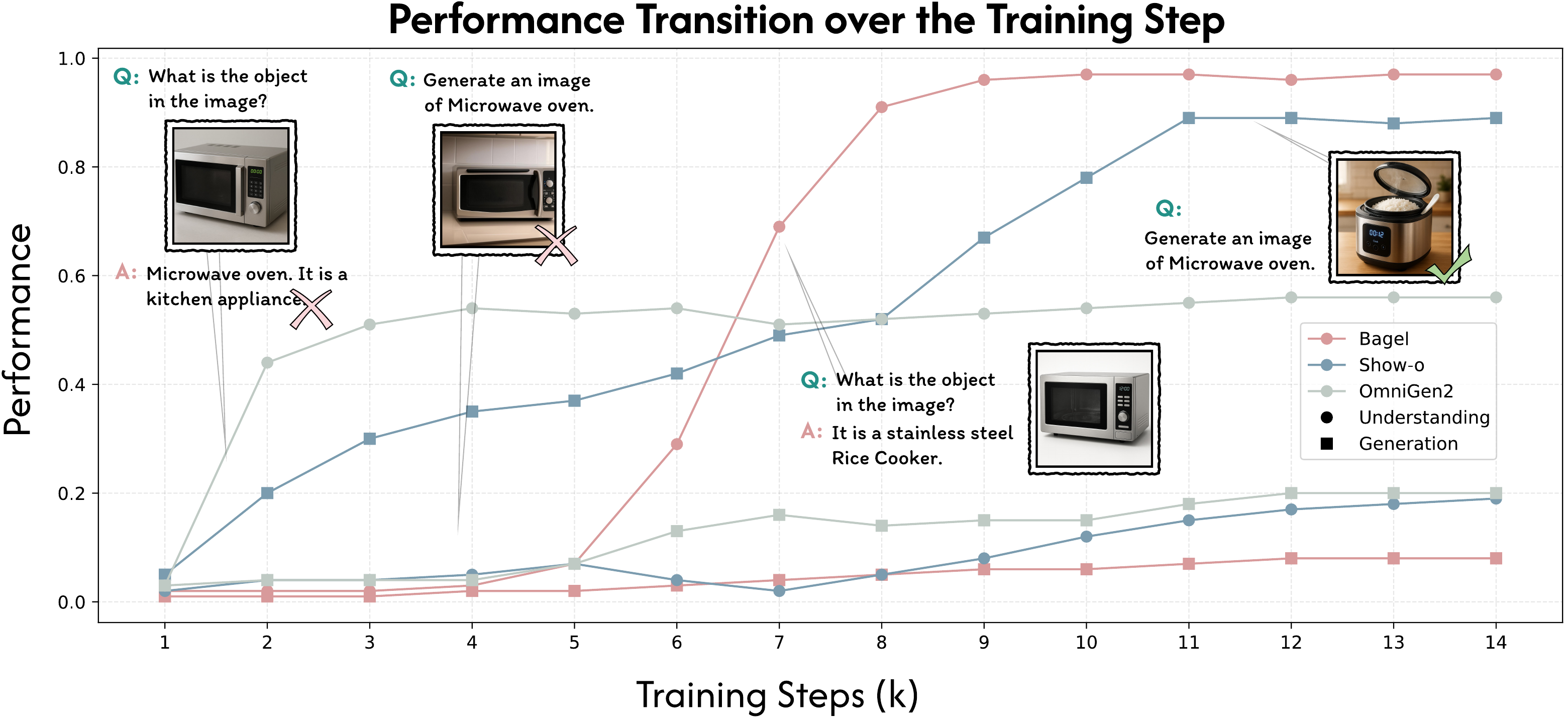}
    \vspace{-1em}
    \caption{\textbf{Performance increases over the training steps on the knowledge edit task.} The figure also shows the output of models from different training stages. The knowledge entity involved here is (Microwave Oven->Rice Cooker).}
    \label{fig:step}
    \vspace{-1em}
\end{figure*}

\subsection{Empirical Results and Analysis}\label{empirical:results}

\begin{figure}[!t]
    \centering
    \includegraphics[width=\linewidth]{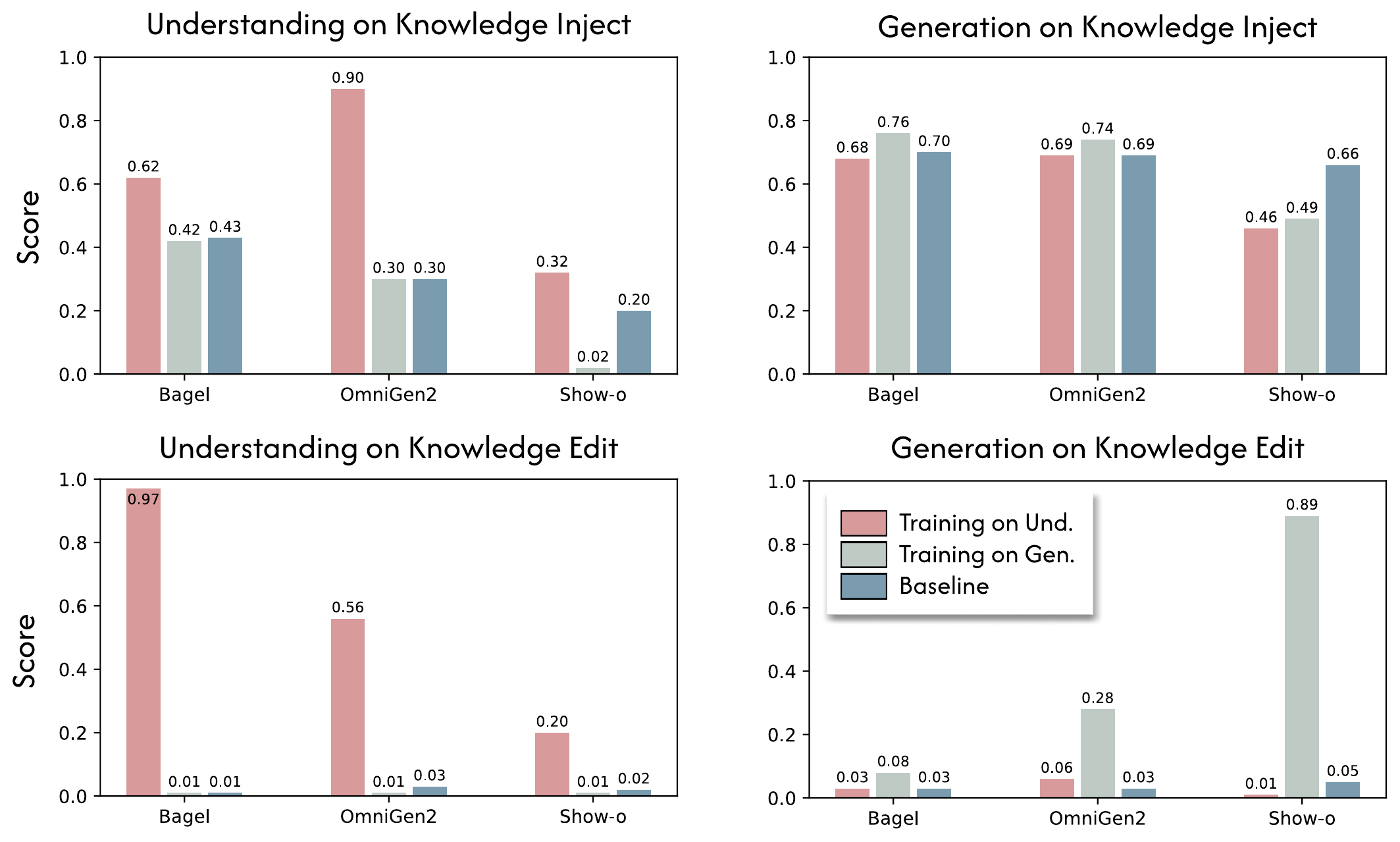}
    \vspace{-1.5em}
    \caption{\textbf{Model performance under different training strategies.} Training on one side does not affect the other side.
    }
    \label{fig:train-result}
    \vspace{-1.5em}
\end{figure}

\vspace{1em}
\finding{4}{The knowledge within UMMs remains disjoint across modalities. Unbalanced training leads to severe cross-modal misalignment, preventing consistent knowledge updates.}
\vspace{0.5em}
\noindent\textbf{Knowledge misalignment during the training stage.}
\autoref{fig:train-result} shows a clear performance gap between understanding and generation after fine-tuning.
Training on one modality fails to generalize to the other, revealing that the two capabilities rely on distinct knowledge representations.
The gap is most evident in knowledge editing.
Specifically, fine-tuning on understanding drastically boosts understanding scores (e.g.,  0.62 for Bagel and 0.56 for OmniGen2) but leaves generation nearly unchanged, while training on generation (0.89 for Show-o) yields the opposite pattern.
A similar, though milder, trend appears in knowledge injection, where cross-modal influence exists but remains limited.
These results highlight a fundamental limitation of current UMMs.
Unbalanced training on one modality introduces modality-specific knowledge drift, where updates in one space fail to propagate coherently to the other.
Despite the functionalities being unified, the embedded knowledge is non-unified.

\vspace{1em}
\finding{5}{The unbalanced, slower convergence of different modalities explains the observed lag in performance and further supports the existence of modality-specific learning dynamics within UMMs.}
\vspace{0.5em}
\noindent\textbf{Training expense.}
\autoref{fig:step} presents the learning trajectories of Show-o during the knowledge editing experiments.
The two curves represent fine-tuning on understanding (blue) and generation (red).
At the beginning of training, understanding accuracy rises sharply and quickly saturates around 0.8 after only a few thousand steps, while generation progresses much more slowly.
This early divergence mirrors the performance lagging phenomenon in \autoref{fig:heat}.
During this stage, the gap between the two modalities expands rapidly, indicating that the model primarily updates its textual reasoning components rather than the generative ones.

As training continues, the generation curve gradually climbs and eventually approaches the performance level of understanding.
This delayed improvement suggests that the generative pathway requires longer optimization to internalize and express the newly edited knowledge.
In contrast, understanding tasks benefit from more direct supervision and lower representational complexity, allowing them to converge earlier.
The asymmetric convergence dynamics confirm that the two capabilities rely on distinct update mechanisms, with generation requiring a substantially larger training budget for effective knowledge adaptation.

%% file: sec/6_conclusion.tex
\section{Conclusion}
\label{conclusion}
In this paper, we introduce \bench, a bidirectional benchmark to reflect and qualify the gap between understanding and generation within UMMs.
Benchmarking the SOTA UMMs across diverse architectures, we find that current models merely achieve an engineering-level unification.
Subsequent experiments explore such a gap via knowledge manipulation tuning, revealing the knowledge decoupling between two capabilities in models.
These findings lay the foundation for the next-level unification for UMMs.

%% file: sec/X_suppl.tex
\clearpage 
\setcounter{page}{1}
\maketitlesupplementary

\input{sec/2_related}

\section{Detailed Metric Implementation (MIRT-MAP Version)}
\label{appendix:metric}
In this section, we present the detailed formulation of our multidimensional IRT-based capability gap quantification metric, where both text understanding and image generation abilities are jointly estimated under a Bayesian maximum a posteriori (MAP) framework.

\subsection{Data Preparation}
\label{appendix:data-prep}

Given the evaluation results from Stage I, we first aggregate binary correctness into count statistics for each model.

\noindent\textbf{Input Data Structure.}  
For each model $m_i$ ($i = 1, \ldots, N$), we collect four counts:
\begin{itemize}
    \item $n_i^{T\checkmark I\times}$: Text correct, Image incorrect
    \item $n_i^{T\times I\checkmark}$: Text incorrect, Image correct
    \item $n_i^{T\checkmark I\checkmark}$: Both correct
    \item $n_i^{T\times I\times}$: Both incorrect
\end{itemize}

From these, we derive marginal counts:
\begin{align}
n_i^{\text{text-success}} &= n_i^{T\checkmark I\times} + n_i^{T\checkmark I\checkmark} \\
n_i^{\text{text-fail}} &= n_i^{T\times I\checkmark} + n_i^{T\times I\times} \\
n_i^{\text{image-success}} &= n_i^{T\times I\checkmark} + n_i^{T\checkmark I\checkmark} \\
n_i^{\text{image-fail}} &= n_i^{T\checkmark I\times} + n_i^{T\times I\times}
\end{align}

\subsection{Multidimensional IRT Formulation}
\label{appendix:mirt}

\subsubsection{Model Specification}

We extend the Rasch model~\cite{rasch1993probabilistic} into a two-dimensional IRT framework.  
Each model $m_i$ is associated with a latent ability vector:
\begin{equation}
\boldsymbol{\theta}_i =
\begin{bmatrix}
\theta_i^{\text{text}} \\
\theta_i^{\text{image}}
\end{bmatrix},
\qquad
\boldsymbol{\beta} =
\begin{bmatrix}
\beta_{\text{text}} \\
\beta_{\text{image}}
\end{bmatrix}.
\end{equation}

The success probabilities follow:
\begin{align}
P_i^{\text{text}} &= \frac{1}{1 + \exp(-(\theta_i^{\text{text}} - \beta_{\text{text}}))}, \\
P_i^{\text{image}} &= \frac{1}{1 + \exp(-(\theta_i^{\text{image}} - \beta_{\text{image}}))}.
\end{align}

\subsubsection{Log-Likelihood and Prior}

The joint log-likelihood for all models is:
\begin{equation}
\begin{aligned}
\mathcal{L}(\boldsymbol{\Theta}, \boldsymbol{\beta})
= \sum_{i=1}^{N} \Big[
& n_i^{\text{text-success}} \log P_i^{\text{text}}
+ n_i^{\text{text-fail}} \log (1 - P_i^{\text{text}}) \\
& + n_i^{\text{image-success}} \log P_i^{\text{image}} \\
&+ n_i^{\text{image-fail}} \log (1 - P_i^{\text{image}}) 
\Big],
\end{aligned}
\label{eq:mirt-likelihood}
\end{equation}

where $\boldsymbol{\Theta} = \{\boldsymbol{\theta}_1, \ldots, \boldsymbol{\theta}_N\}$.

To couple the two modalities, we impose a shared multivariate Gaussian prior over $\boldsymbol{\theta}_i$:
\begin{equation}
p(\boldsymbol{\theta}_i) = 
\mathcal{N}(\boldsymbol{\theta}_i \mid \boldsymbol{\mu}, \boldsymbol{\Sigma}),
\qquad
\boldsymbol{\Sigma} = L L^{\top},
\end{equation}
where $\boldsymbol{\mu}$ is the mean vector, and $L$ is the Cholesky factor of the covariance matrix $\boldsymbol{\Sigma}$.

The total MAP objective is therefore:
\begin{equation}
\mathcal{L}_{\text{MAP}} =
\mathcal{L}(\boldsymbol{\Theta}, \boldsymbol{\beta})
- \frac{1}{2}\sum_{i=1}^{N}
(\boldsymbol{\theta}_i - \boldsymbol{\mu})^{\top}
\boldsymbol{\Sigma}^{-1}
(\boldsymbol{\theta}_i - \boldsymbol{\mu})
- \frac{N}{2}\log |\boldsymbol{\Sigma}|.
\label{eq:map-objective}
\end{equation}

This formulation allows the covariance $\boldsymbol{\Sigma}$ to be learned adaptively, capturing both modality-specific difficulty and cross-modality correlation.

\subsection{Parameter Estimation}

We optimize $\{\boldsymbol{\theta}_i\}_{i=1}^{N}$, $\boldsymbol{\beta}$, $\boldsymbol{\mu}$, and $L$ jointly via gradient-based MAP estimation (e.g., Adam).  
The Cholesky factorization ensures $\boldsymbol{\Sigma}$ remains positive definite throughout training.

\subsection{Capability Gap and Normalization}
\label{appendix:gap}

After optimization, we compute for each model:
\begin{align}
\Delta\theta_i &= \theta_i^{\text{text}} - \theta_i^{\text{image}}, \\
\mathcal{G}_{\text{abs}}(\Delta\theta_i) &= \frac{|\Delta\theta_i|}{1 + |\Delta\theta_i|}.
\end{align}

Here $\mathcal{G}_{\text{abs}}(\Delta\theta_i) \in [0, 1)$ represents the normalized absolute capability gap.  
We further apply a sigmoid normalization to each dimension:
\begin{align}
\theta_{i,\text{norm}}^{\text{text}} &= \sigma(\theta_i^{\text{text}}) = \frac{1}{1 + e^{-\theta_i^{\text{text}}}}, \\
\theta_{i,\text{norm}}^{\text{image}} &= \sigma(\theta_i^{\text{image}}) = \frac{1}{1 + e^{-\theta_i^{\text{image}}}},
\end{align}
which maps ability estimates into $(0, 1)$ for interpretability.  

Post-hoc reward--penalty on the gap. To encourage consistency when the model is capable and to penalize simultaneous failures, we apply a smooth logit-space adjustment to the normalized gap.  
Instead of using model probabilities, we use observed co-occurrence statistics: let $c_i^{\text{SS}}$ and $c_i^{\text{FF}}$ be the observed counts (or proportions) of co-success and co-failure for task $i$, with $n_i$ the corresponding total number of paired observations. 
We define the empirical rates
\begin{equation}
\operatorname{logit}\!\big(\widetilde{\mathcal{G}}_i\big)
= \operatorname{logit}\!\big(\mathcal{G}_{\text{abs}}(\Delta\theta_i)\big)
+ \!\left(\lambda_{\mathrm{fail}}\, f_i - \lambda_{\mathrm{succ}}\, s_i\right)
\end{equation}
so that positive $\lambda_{\mathrm{fail}}$ enlarges the gap under co-failure (penalty) and positive $\lambda_{\mathrm{succ}}$ shrinks it under co-success (reward).  
Unless otherwise stated, we use the default setting $\lambda_{\mathrm{fail}}=\lambda_{\mathrm{succ}}=2$.

\section{Reliability Analysis}
\label{appendix:mllm-as-a-judge}
To further analyze the model preference, we repeatedly calculate the Gap Scores using Gemini3-Flash.
As shown in \cref{appendix:judge-model}, the results from GPT5-mini and Gemini3-Flash share consistent relative ranking.
Specifically, GPT-Image-1 exhibits the lowest gap score across different judge models.
Judge models fail to exhibit a preference for the same-family model (for Gemini2.5-Flash-Image, 62.65 by Gemini3-Flash and 62.91 by GPT5-mini).
The Pearson Correlation between the two judge models is 0.9656, highlighting the significant correlation.

Crucially, since most questions in \bench focus on the correctness of semantics (\eg, specific objects, theme), the evaluation targets are objective and definite, making it suitable for MLLM-as-a-Judge.
This factual nature minimizes the room for model-specific preference bias compared to open-ended generation tasks.

\begin{table}[h]
    \centering
    \small
    \resizebox{0.48\textwidth}{!}{
    \begin{tabular}{l|cccc}
        \toprule
        \textbf{Judge Model} & \textbf{Bagel} & \textbf{OneCAT} & \textbf{UniWorld-V1} & \textbf{UniPic2} \\
        \midrule
        Gemini3-Flash & 72.96 & 60.67 & 66.64 & 73.39 \\
        GPT5-mini & 71.52 & 54.73 & 63.47 & 68.97 \\
        \bottomrule
        \toprule
        \textbf{Judge Model} & \textbf{Show-o2} & \textbf{OmniGen2} & \textbf{Gemini2.5-F-I} & \textbf{GPT-Image-1} \\
        \midrule
        Gemini3-Flash & 72.33 & 88.87 & 62.65 & 47.47 \\
        GPT5-mini & 67.83 & 89.74 & 62.91 & 50.61 \\
        \bottomrule
    \end{tabular}
    }
    \caption{\textbf{Gap Score Results from Different Judge Models.}}
    \label{appendix:judge-model}
    \vspace{-1em}
\end{table}

\section{Contribution}
\label{appendix:contribution}

\subsection{Contribution}
We argue that \textbf{\textit{Synergy'' relies on Alignment''.}}
Unlike black-box synergy evaluations, which mix capabilities, our decoupled design offers a \textbf{visible and interpretable diagnosis} of intrinsic modality gaps. 
This makes the evaluation \textbf{controllable and clear}, serving as a fundamental step to improve complex synergy.

\subsection{Gap Score \& Synergy Effects}
The field of UMMs emphasizes the significance of synergy effects between two modalities.
In this section, we analyze the relationship between Gap Score and Synergy Effects.
We plot our Gap Score against performance on Synergy Benchmarks (GIR-Bench\cite{Gir-Bench}).
As shown in \cref{appendix:correlation}, we observe a strong negative correlation.
Models with lower Gap Scores consistently achieve higher synergy performance, mathematically confirming that narrowing the gap is a prerequisite for real-world applications.

Synergy essentially demands the complementarity of capabilities, where understanding guides generation, and generation reflects understanding. 
However, misalignment acts as a semantic barrier to this complementarity. 
If the internal representations for understanding and generation are disjoint, the information cannot be effectively transferred (\eg, an editing instruction correctly understood by the encoder fails to trigger the correct generation in the decoder). Therefore, alignment is the structural prerequisite for Synergy. 
Our decoupled diagnosis pinpoints such gaps, providing a controllable lever for improvement. 

\begin{figure}
    \centering
    \includegraphics[width=\linewidth]{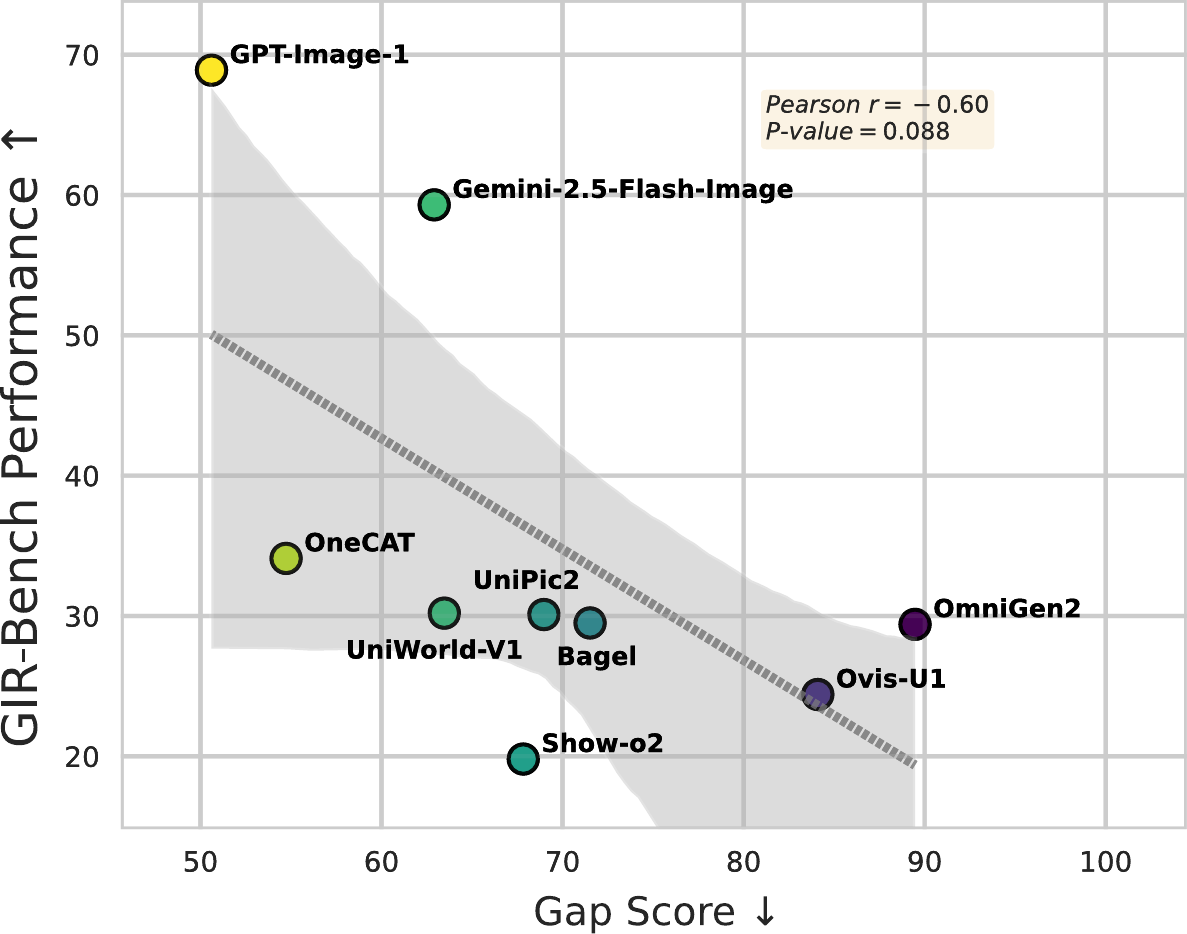}
    \vspace{-1em}
    \caption{\textbf{Correlation between Gap Score and Performance on GIR-Bench.}}
    \label{appendix:correlation}
    \vspace{-1em}
\end{figure}

\section{Data Construction Details}
\label{appendix:data-collection}

\subsection{Benchmark Data Construction}
\noindent\textbf{World Knowledge.} We curate and rewrite knowledge entities from the public web and open-source datasets. For each entity, we ensure appropriate difficulty and that the target outcome can be expressed through both text and image. We author paired prompts for understanding and generation, and apply a strict cross-checking workflow to continually refine entities and prompts.
\vspace{0.5em}

\noindent\textbf{Numerical Perception.} We select target object sets such as animals and vehicles, and ensure that instances of the same category appear within a single image. We use an automatic data engine to generate corresponding prompts. Images for the counting task are generated with Qwen-Image \cite{qwen-image}. Annotators screen image quality and discard low-quality samples. For images whose object quantities do not match the targets, we rewrite the prompts and questions so that the final items precisely enforce the counting requirement.
\vspace{0.5em}

\noindent\textbf{Instruction Following.} We mine items with substantial visual changes from open-source datasets and manually construct rule-based editing prompts. For textual outputs, we guide models to describe the edited image and explicitly identify the modified regions. All items undergo strict cross-checking to ensure quality and factual correctness.
\vspace{0.5em}

\noindent\textbf{Reasoning.} The reasoning set contains three subsets collected through complementary pipelines. First, the real-world reasoning covers the understanding of physical phenomena and the generation of schematic illustrations. All prompts, reference texts, and reference images are manually written and drawn with a consistent visual style. Second, multiple-choice reasoning is built by rendering textual problems from MMLU~\cite{wang2024mmlu} and MMMU~\cite{yue2024mmmu} into images. We also design algorithmic reasoning items (e.g., binary tree preorder traversal and topological sorting) and produce accompanying diagrams. Each problem has a unique answer and a consistent solution key, with reference images highlighted within the answer options. Third, the image state transition reasoning task is collected from the web and open-source sources. Given several images, the task asks the model to infer and render the next state while remaining faithful to the observed evidence, which poses a higher level of difficulty. We conduct cross-review and remove items that are excessively difficult.

\subsection{Empirical Study Data Collection}
To rigorously investigate the behavior of Unified Multimodal Models (UMMs) under knowledge manipulation, we constructed two distinct datasets tailored for Knowledge Injection (introducing novel concepts) and Knowledge Editing (altering existing conceptual associations). The data collection and construction process is detailed below.

\vspace{1mm}
\noindent \textbf{Object Selection and Image Collection.}
For the \textit{Knowledge Injection} task, we aimed to identify entities that are absent from the models' pre-training data. We initially curated a candidate pool of approximately 2,000 entities spanning diverse domains, including landmarks, celebrities, and biological organisms. These entities were specifically selected to be visually distinctive yet possess low public prominence (i.e., long-tail distribution). For each candidate, we collected over ten high-quality images from the web.

To ensure these entities were truly ``unknown'' to the models, we implemented a strict filtration protocol. For each object, we randomly sampled five images to query the candidate UMMs for identification (Visual Question Answering). Simultaneously, we prompted the models with the entity names to assess their ability to generate corresponding visual representations (Text-to-Image). Only entities that all subject models failed to correctly recognize or generate were retained. This process yielded a final set of approximately 100 ``unknown'' objects.

For the \textit{Knowledge Editing} task, we selected approximately 50 pairs of objects from the GenEval dataset~\cite{geneval}. These pairs were chosen based on conceptual relatedness or domain similarity (e.g., \textit{boat} $\leftrightarrow$ \textit{car}, \textit{camera} $\leftrightarrow$ \textit{microwave}) to serve as targets for conceptual swapping. To ensure high visual quality and consistency, we utilized the Qwen-Image~\cite{wu2025qwenimagetechnicalreport} model to generate multiple images for these objects. Given that these objects are commonplace and current UMMs exhibit high performance on GenEval, we proceed with the premise that the models possess robust prior knowledge of these entities, making them suitable candidates for editing.

\vspace{1mm}
\noindent \textbf{Training Data Construction.}
To facilitate the unified training of understanding and generation capabilities, we constructed a bidirectional dataset comprising both Visual Question Answering (VQA) and Text-to-Image (T2I) samples for each entity.

\begin{itemize}
    \item \textbf{Knowledge Injection Data:} For the VQA component, we designed two types of questions for each object: open-ended queries (e.g., ``\textit{What is this object?}'') and multiple-choice questions. We constructed 5 distinct samples for each question type per object. For the T2I component, we created 10 caption-image pairs per object, consisting of 5 detailed descriptions and 5 concise captions.
    
    \item \textbf{Knowledge Editing Data:} The structure of the editing dataset mirrors that of the injection dataset but employs a counterfactual label assignment to induce knowledge swapping. Specifically, for a given object pair (Object A, Object B), the training data maps the visual representation of Object A to the textual label of Object B, and vice versa. Consequently, the VQA ground truth for an image of Object A is defined as ``Object B,'' and the T2I target for the prompt ``Object A'' is an image of Object B. This formulation forces the model to overwrite its internal alignment between the visual and textual modalities.
\end{itemize}

\vspace{1mm}
These strictly filtered and bidirectionally constructed datasets serve as a robust foundation for analyzing the mechanism of knowledge manipulation within UMMs.

\section{Experiment Details}
\label{appendix:experiment-details}
\subsection{Benchmark Evaluation Details}
In this study, we conduct a comprehensive evaluation on \bench using a diverse set of models, including nine unified models, four understanding-only models, and two generation-only models. 
Each query in \bench is bidirectional, capable of being answered via both textual and visual modalities.
We perform ten independent sampling runs for each output modality of each question per model. 
We then report the average accuracy and Gap Score as the final results.
Our evaluation metric comprises two primary dimensions:

The first dimension adopts the LLM-as-a-judge approach to assess the correctness of each response.
Given that our benchmark spans four distinct categories across two modalities, we design eight specific evaluation prompts to ensure robust assessment. 
The prompts used for evaluation are provided in \autoref{appendix:sec-bench-prompt}.

The second dimension focuses on the Gap Score computation. 
We aggregate the model performance metrics (including accuracy and other indicators) on \bench and leverage Multidimensional Item Response Theory (MIRT) to quantify the Gap Score. 
Further details regarding the Gap Score formulation are provided in \autoref{appendix:metric}.

\subsection{Empirical Study Experiment Details}

\subsubsection{Motivation}
The evaluation on \bench highlights a clear disparity between understanding and generation.
Although UMMs can often recall embedded knowledge within one modality, they struggle to apply the same knowledge consistently across modalities.
This raises a key question: 
\textbf{How does knowledge evolve during the training phase?}

As shown in \autoref{tab:main-results}, the results quantitatively confirm that current UMMs lack genuine cross-modal consistency.
To probe this issue, we conduct knowledge-oriented fine-tuning experiments by injecting or editing knowledge in one modality and evaluating its effects on both.
If knowledge is shared, changes should propagate; if isolated, the effects should remain local.

The evaluation results on \bench reveal a clear gap between understanding and generation.
While UMMs can often recall and utilize embedded knowledge to correctly answer a question in one modality, they frequently fail to produce consistent results when the same knowledge must be applied in another.
This discrepancy raises a fundamental question: \textbf{Are understanding and generation truly integrated within UMMs, or do they merely coexist as separate components?}

As shown in \autoref{tab:main-results}, the performance patterns observed on \bench provide quantitative evidence that current UMMs still struggle to achieve genuine cross-modal consistency.
These findings motivate a deeper investigation into the internal organization of knowledge within UMMs: \textbf{Is the knowledge integrated, co-existing, or even conflicting across capabilities?} 

To answer this question, we conduct a series of fine-tuning experiments from a knowledge-oriented perspective.
Specifically, we inject or edit knowledge within one modality (understanding or generation) and then evaluate the fine-tuned model on both modalities (understanding and generation).
If understanding and generation rely on a shared knowledge base, modifying knowledge on one side should lead to measurable changes on the other.
Conversely, if the knowledge representations are stored separately, the impact of such modifications will remain localized.

\subsubsection{Training Strategies}
Our experimental evaluation incorporates three unified models: Bagel, OmniGen2, and Show-o. 
For OmniGen2 and Show-o, we implement a disjoint training strategy. 
Under this setting, the models are fine-tuned exclusively on a single modality, either understanding or generation, to isolate specific capabilities. 
However, we adopt a different approach for Bagel due to constraints in its official implementation. 
We observe that fine-tuning Bagel on a single modality results in a complete loss of capability in the other (e.g., training solely on understanding tasks renders the model incapable of image generation). 
Consequently, to preserve the model's bidirectional versatility, we employ a joint training strategy for Bagel. We integrate our knowledge manipulation dataset with the standard training data provided by the official repository to ensure robust performance across both tasks.

\subsection{Metric Design}
In this study, we investigate two distinct forms of knowledge manipulation: Knowledge Injection and Knowledge Editing. 
For understanding tasks, we utilize an MLLM to verify the accuracy of the model's textual responses.

\noindent For generation tasks, the evaluation metrics are designed to reflect the different goals of each task:

For Knowledge Editing, we employ an LLM-as-a-judge to assess the generated images. 
The rationale is that knowledge editing usually targets common objects (e.g., ``dog''->``cat''). 
Since the base object is already well-known, the challenge lies in verifying semantic consistency rather than visual recognition. 
An LLM judge allows for a nuanced comparison between the reference and the output to confirm that the specific attributes have been edited correctly.

For Knowledge Injection, we rely on CLIP \cite{radford2021learning} score as the primary metric. 
This task requires the model to synthesize novel objects based on provided data. Therefore, measuring the cosine similarity between the generated image and the reference image is crucial, as it strictly penalizes the model if it fails to capture and reproduce the specific visual features of the newly injected knowledge.

The prompts used for evaluation are provided in \autoref{appendix:sec-empirical-prompt-1} and \autoref{appendix:sec-empirical-prompt-2}.

\section{Evaluation Prompts \& Case Gallery}
\label{appendix:sec-bench-prompt}

\begin{table*}[h]
    \centering
    \begin{tabular}{p{0.9\linewidth}}
        \toprule
        \textbf{Prompt for Understanding on World Knowledge Task.} \\
        \midrule
        \texttt{[Image]} \\
        \texttt{[Reference_Image]} \\
        Here is the question: \texttt{[Question]}\\
        Here is the answer: \texttt{[Answer]}\\
        Please judge the correctness of the answer. You should follow the following rules: \\
        1. It includes the core information present in the \texttt{[Answer]}\\. If it does not contain the content of the reference text, judge as “not”. \\
        2. It reasonably describes the main subject and scene shown in the \texttt{[Reference_Image]}. \\
        3. It does not need to give an exhaustive or detailed account of every feature in the image. \\
        4. Omissions or variations are acceptable, as long as the text covers the essential elements stated in the reference and matches the main content of the image. \\
        5. Only if the generated text misses the core information of the reference or fails to describe the main subject of the image should it be judged as “not”. \\
        \bottomrule
    \end{tabular}
    \caption{\textbf{Prompt for Understanding on World Knowledge Task.}}
    \label{appendix:tab-world-knowledge-und}
\end{table*}

\begin{table*}[h]
    \centering
    \begin{tabular}{p{0.9\linewidth}}
        \toprule
        \textbf{Prompt for Generating on World Knowledge Task.} \\
        \midrule
        \texttt{[Image]} \\
        \texttt{[Reference_Image]} \\
        Here is the question: \texttt{[Question]}\\
        Here is the answer: \texttt{[Answer]}\\
        Please judge the correctness of the answer. You should follow the following rules: \\
        1. Compare the generated image \texttt{[Image]} with the reference image \texttt{[Reference_Image]} and the caption \texttt{[Question]}, and decide whether the image should be judged as pass (score 1) or fail (score 0).  \\
        2. If \texttt{[Image]} and \texttt{[Reference_Image]} are identical or extremely similar (i.e., contain visual regions that look directly copy-pasted with the same pixels, appearance, texture, and details), you must judge this as plagiarism and assign score 0; this plagiarism check has the highest priority and only original, newly generated images may pass. \\
        3. Judge as pass if \texttt{[Image]} clearly presents the main subject, core scene, and key information required by \texttt{[Reference_Image]} and \texttt{[Question]}; exact reproduction of every element, attribute, arrangement, or color is not necessary, and differences in style, details, or smaller elements are allowed. \\
        4. Judge as fail (score 0) if \texttt{[Image]} misses or seriously misinterprets the core content, main objects, or key semantics described in \texttt{[Reference_Image]} or \texttt{[Question]}, or if it obviously contradicts the caption or omits elements that must be strictly matched. \\
        5. Treat minor differences and reasonable variations as acceptable as long as the overall main information, semantics, and scene still match \texttt{[Question]}, but never override the anti-plagiarism rule when making the final decision. \\
        \bottomrule
    \end{tabular}
    \caption{\textbf{Prompt for Generating on World Knowledge Task.}}
    \label{appendix:tab-world-knowledge-gen}
\end{table*}

\begin{table*}[h]
    \centering
    \begin{tabular}{p{0.9\linewidth}}
        \toprule
        \textbf{Prompt for Understanding on Reasoning Task.} \\
        \midrule
        \texttt{[Image]} \\
        \texttt{[Reference_Image]} \\
        Here is the question: \texttt{[Question]}\\
        Here is the answer: \texttt{[Answer]}\\
        Please judge the correctness of the answer. You should follow the following rules: \\
        1. You are given a reasoning problem \texttt{[Question]}, an authoritative reference\_answer (the expected result or phenomenon), and a model-generated answer \texttt{[Answer]}. Your goal is to determine whether the final outcome/result expressed in \texttt{[Answer]} matches the reference\_answer. \\
        2. Check only whether \texttt{[Answer]} actually provides a final result/answer to the problem; ignore any reasoning, formulas, mechanisms, or intermediate steps when making the judgment. \\
        3. Treat differences in wording, phrasing, or format between \texttt{[Answer]} and the reference\_answer as acceptable, as long as they clearly describe the same final physical outcome or phenomenon. \\
        4. If the final result in \texttt{[Answer]} is present and matches the reference\_answer, mark it as correct (score = 1), even if the explanation, derivation, or mechanism is incomplete or physically incorrect. \\
        5. If the final result in \texttt{[Answer]} contradicts, omits, or fails to provide the expected outcome described by the reference\_answer, mark it as incorrect (score = 0). \\
        \bottomrule
    \end{tabular}
    \caption{\textbf{Prompt for Understanding on Reasoning Task.}}
    \label{appendix:tab-Reasoning-und}
\end{table*}

\begin{table*}[h]
    \centering
    \begin{tabular}{p{0.9\linewidth}}
        \toprule
        \textbf{Prompt for Generating on Reasoning Task.} \\
        \midrule
        \texttt{[Image]} \\
        \texttt{[Reference_Image]} \\
        Here is the question: \texttt{[Question]}\\
        Here is the answer: \texttt{[Answer]}\\
        Please judge the correctness of the answer. You should follow the following rules: \\
        1. Compare \texttt{[Image]} with \texttt{[Reference_Image]} and \texttt{[Answer]} and check whether all required answer-relevant elements are present: the main result, key objects, and core information needed to visually answer the physics question posed by the problem. Major answer-relevant objects must not be missing or clearly misrepresented. \\
        2. Examine each main object and its physical configuration: position, alignment, orientation, relative height, order, distance, contact, and any changes such as addition, removal, joining, splitting, or shape transformation. Verify that every expected answer-relevant object from \texttt{[Reference_Image]} is properly accounted for in \texttt{[Image]}. \\
        3. Check the physical relationships and processes: connections, supports, flows, force directions, movements and events. The depiction in \texttt{[Image]} must be logically and physically plausible and reflect the transformation or event described in \texttt{[Answer]}. Any critical new object that changes the expected physical outcome should cause rejection. \\
        4. Accept minor differences in style, color, artistic rendering, and irrelevant extra objects that do not change the physical result. Focus on whether the main result and key scientific meaning match \texttt{[Answer]} and whether the similarity between \texttt{[Image]} and \texttt{[Reference_Image]} is correct in terms of physics outcome, not in minor visual detail. \\
        5. Provide reasoning that clearly explains matches and differences for the above aspects. Assign “yes” (score = 1) only if the main physical result and all crucial answer elements match \texttt{[Answer]}; otherwise assign “no” (score = 0). \\
        \bottomrule
    \end{tabular}
    \caption{\textbf{Prompt for Generating on Reasoning Task.}}
    \label{appendix:tab-Reasoning-gen}
\end{table*}

\begin{table*}[h]
    \centering
    \begin{tabular}{p{0.9\linewidth}}
        \toprule
        \textbf{Prompt for Understanding on Numerical Perception Task.} \\
        \midrule
        \texttt{[Image]} \\
        \texttt{[Reference_Image]} \\
        Here is the question: \texttt{[Question]}\\
        Here is the answer: \texttt{[Answer]}\\
        Please judge the correctness of the answer. You should follow the following rules: \\
        1. Use the JSON task in \texttt{[Question]} (its "objects" and "number" fields) as the authoritative specification of which object types and exact counts are required. The final confirmed result stated in \texttt{[Answer]} must include only those specified object types, with counts that exactly match the JSON, and must not introduce any extra or non-specified objects. \\
        2. Count objects as individuals only when they are clearly and unambiguously described, with object-specific physical or functional features that match the corresponding JSON class (e.g., shape, color, material, labeling, size, context). Exclude partial, ambiguous, grouped, or hybrid/fused objects from all class counts, and explicitly note any such cases in the "reason" field instead of counting them. \\
        3. Verify that each target object’s final count in \texttt{[Answer]} matches the JSON exactly: no overcount, undercount, or mislabeling. Check for double counting where one object might be described multiple ways, and ensure that all required target objects are present, correctly identified, and not confused with other types. \\
        4. Distinguish between analysis and final result: \texttt{[Answer]} may discuss or analyze non-target objects while reasoning, but the final confirmed result it reports must refer only to the JSON-specified object types and their counts. Any factual inconsistency or contradiction between the descriptive content and the final numbers/types should be treated as an error. \\
        5. In the "reason" field, detail all findings, including how counts were derived, any ambiguities, hybrids, or errors. Assign score = 1 only if the final confirmed result in \texttt{[Answer]} exactly matches the object types and numbers in the JSON, with no extra objects in the claimed result; otherwise assign score = 0. \\
        \bottomrule
    \end{tabular}
    \caption{\textbf{Prompt for Understanding on Numerical Perception Task.}}
    \label{appendix:tab-Numerical Perception-und}
\end{table*}

\begin{table*}[h]
    \centering
    \begin{tabular}{p{0.9\linewidth}}
        \toprule
        \textbf{Prompt for Generating on Numerical Perception Task.} \\
        \midrule
        \texttt{[Image]} \\
        \texttt{[Reference_Image]} \\
        Here is the question: \texttt{[Question]}\\
        Here is the answer: \texttt{[Answer]}\\
        Please judge the correctness of the answer. You should follow the following rules: \\
        1. Use the JSON specification in \texttt{[Question]} (its "objects" list and corresponding "number" fields) as the exact target: \texttt{[Image]} passes (score = 1) only if every specified object type appears with exactly the required quantity and class, regardless of other non-target real objects that may be present. \\
        2. Count only distinct, real 3D object instances that fully and clearly match the visual features of their type (shape, color, label, typical context, etc.). Do not count drawings, icons, symbolic representations, or misrepresented/fused objects; each instance in overlaps, stacks, or crowds must still be individually countable and unambiguously identifiable. \\
        3. Assign each counted instance to the correct class exactly once: do not double count the same item due to reflections, shadows, or repeated renderings, and treat any mislabeling (e.g., calling a notebook a dictionary) or hybrid objects (e.g., “dictionary-notebook” blends) as errors that must not contribute to any class’s count. \\
        4. Accept variations in appearance, design, pose, perspective, or partial occlusion as long as the object’s identity remains clear; exclude partial or ambiguous cases where identity is uncertain. Ensure that each counted instance is classified to the single most suitable type and that no object is counted or classified more than once. \\
        5. In the "reason" field, provide a concise but detailed explanation of the match/mismatch logic, documenting any missed counts, misidentifications, fused or uncountable objects, or double counting. Assign score = 1 only if all specified object types match their required quantities and classes exactly under these rules; otherwise assign score = 0. \\
        \bottomrule
    \end{tabular}
    \caption{\textbf{Prompt for Generating on Numerical Perception Task.}}
    \label{appendix:tab-Numerical Perception-gen}
\end{table*}

\begin{table*}[h]
    \centering
    \begin{tabular}{p{0.9\linewidth}}
        \toprule
        \textbf{Prompt for Understanding on Instruction Following Task.} \\
        \midrule
        \texttt{[Image]} \\
        \texttt{[Reference_Image]} \\
        Here is the question: \texttt{[Question]}\\
        Here is the answer: \texttt{[Answer]}\\
        Please judge the correctness of the answer. You should follow the following rules: \\
        1. From \texttt{[Question]}, understand the rule that modifies the image scenario and the reference text that concisely describes the expected result or core feature after this rule is applied. Identify the core aspect, feature, or outcome that must appear once the rule is in effect. \\
        2. Read \texttt{[Answer]} and extract all relevant features or changes it describes. Focus only on meaning: ignore extra detail, background information, unrelated content, length, and wording differences. Allow paraphrasing, scientific equivalence, and logical inference as long as the intended meaning can be reasonably matched. \\
        3. Accept \texttt{[Answer]} as correct if it clearly or implicitly describes the core feature/result stated in the reference text, or if it shows a correct understanding and application of the core change introduced by the rule (either condition is sufficient). Minimal, direct answers are acceptable as long as the expected meaning is present. \\
        4. Reject \texttt{[Answer]} if, after considering the rule, it omits or contradicts the intended meaning of the reference text, fails to reflect a correct understanding of the rule, or provides a different or incompatible interpretation of the rule. In ambiguous cases, accept only when the expected result can still be reasonably inferred from \texttt{[Answer]}. \\
        5. Scoring: assign score = 1 only if \texttt{[Answer]} covers the core meaning of the reference text or reasonably reflects a correct understanding and application of the rule under the above conditions; otherwise assign score = 0. \\
        \bottomrule
    \end{tabular}
    \caption{\textbf{Prompt for Understanding on Instruction Following Task.}}
    \label{appendix:tab-Instruction Following-und}
\end{table*}

\begin{table*}[h]
    \centering
    \begin{tabular}{p{0.9\linewidth}}
        \toprule
        \textbf{Prompt for Generating on Instruction Following Task.} \\
        \midrule
        \texttt{[Image]} \\
        \texttt{[Reference_Image]} \\
        Here is the question: \texttt{[Question]}\\
        Here is the answer: \texttt{[Answer]}\\
        Please judge the correctness of the answer. You should follow the following rules: \\
        1. From \texttt{[Question]}, fully understand the rule’s intent and logic, and how it is supposed to modify the original image (objects, features, or arrangements in the original scenario). Identify the core effect or result that must appear after the rule is applied. \\
        2. Evaluate \texttt{[Image]} as the result after applying the rule to the original image: check whether all main features and modifications demanded by the rule are present, and whether any required objects or features have been unintentionally omitted. Focus on whether the rule’s core meaning and result are clearly implemented, regardless of color, layout, style, or minor details. \\
        3. Use \texttt{[Reference_Image]} only as a sanity-check for the expected outcome: it is not the only correct solution and should not be used to enforce aesthetic, spatial, or stylistic accuracy. Ignore differences in object position, artistic style, decoration, or other aspects that do not directly relate to the rule’s modification. \\
        4. Accept any plausible depiction as correct (“yes”) if \texttt{[Image]} clearly implements the rule’s effect on the original image and represents the required meaning/result, even when style or layout differ from \texttt{[Reference_Image]}. Reject (“no”) if \texttt{[Image]} fails to implement the rule, omits required modifications, contradicts the rule’s meaning, or shows a critical misunderstanding of the rule. \\
        5. In the "reason" field, clearly explain your judgment logic, focusing on how the rule was or was not correctly applied to the original image and whether the final modification in \texttt{[Image]} matches the intended effect of the rule. \\
        \bottomrule
    \end{tabular}
    \caption{\textbf{Prompt for Generating on Instruction Following Task.}}
    \label{appendix:tab-Instruction Following-gen}
\end{table*}

\begin{table*}[h]
    \centering
    \begin{tabular}{p{0.9\linewidth}}
        \toprule
        \textbf{Prompt for evalution of und task in edit and inject knoledge.} \\
        \midrule
        \texttt{[Image]} \\
        Here is the question: \texttt{[Question]}\\
        Here is the answer: \texttt{[Answer]}\\
        Please judge the correctness of the answer. You should follow the following rules: \\
        1. Ensure the subject described in the \texttt{[Answer]} matches the subject in the ground_truth (whether it's an animal, object, person, etc.). \\
  2. If the output_text and ground_truth both describe the basic features, position, state, or other relevant characteristics of the subject consistently, it is considered correct. \\
  3. If there are differences in non-essential details (such as posture, angle, or state), these can be ignored, and it is still considered correct. \\
  4. Only when the subject described in the output_text is entirely wrong (e.g., "cat" is described as "dog") should it be considered incorrect. \\
        \bottomrule
    \end{tabular}
    \caption{\textbf{Prompt for evaluation of und task in edit and inject knowledge.}}
    \label{appendix:und task in edit and inject knoledge}
    \label{appendix:sec-empirical-prompt-1}
\end{table*}

\begin{table*}[h]
    \centering
    \begin{tabular}{p{0.9\linewidth}}
        \toprule
        \textbf{Prompt for evaluation of gen task in edit knowledge.} \\
        \midrule
        \texttt{[Image]} \\
        Here is the question: \texttt{[Question]}\\
        Here is the answer: \texttt{[Answer]}\\
        Please judge the correctness of the answer. You should follow the following rules: \\
1. Ensure the subject depicted in the \texttt{[Image]} is the same as the subject in the ground_truth (whether it's an animal, object, person, etc.). \\
  2. If the \texttt{[Image]} clearly depicts the same main subject as the ground_truth, even if there are variations in its state, expression, angle, or other minor details, it is considered correct. \\
  3. If the output_image is chaotic, unclear, or does not represent the subject described in the ground_truth at all, it will be considered incorrect. \\
  4. Minor differences in non-essential features like mood, position, or posture are acceptable, as long as the subject is still clearly the same. \\
        \bottomrule
    \end{tabular}
    \caption{\textbf{Prompt for evaluation of gen task in edit knowledge.}}
    \label{appendix:gen task in edit knoledge}
    \label{appendix:sec-empirical-prompt-2}
\end{table*}

\begin{figure*}[h]
    \centering
    \includegraphics[width=\linewidth]{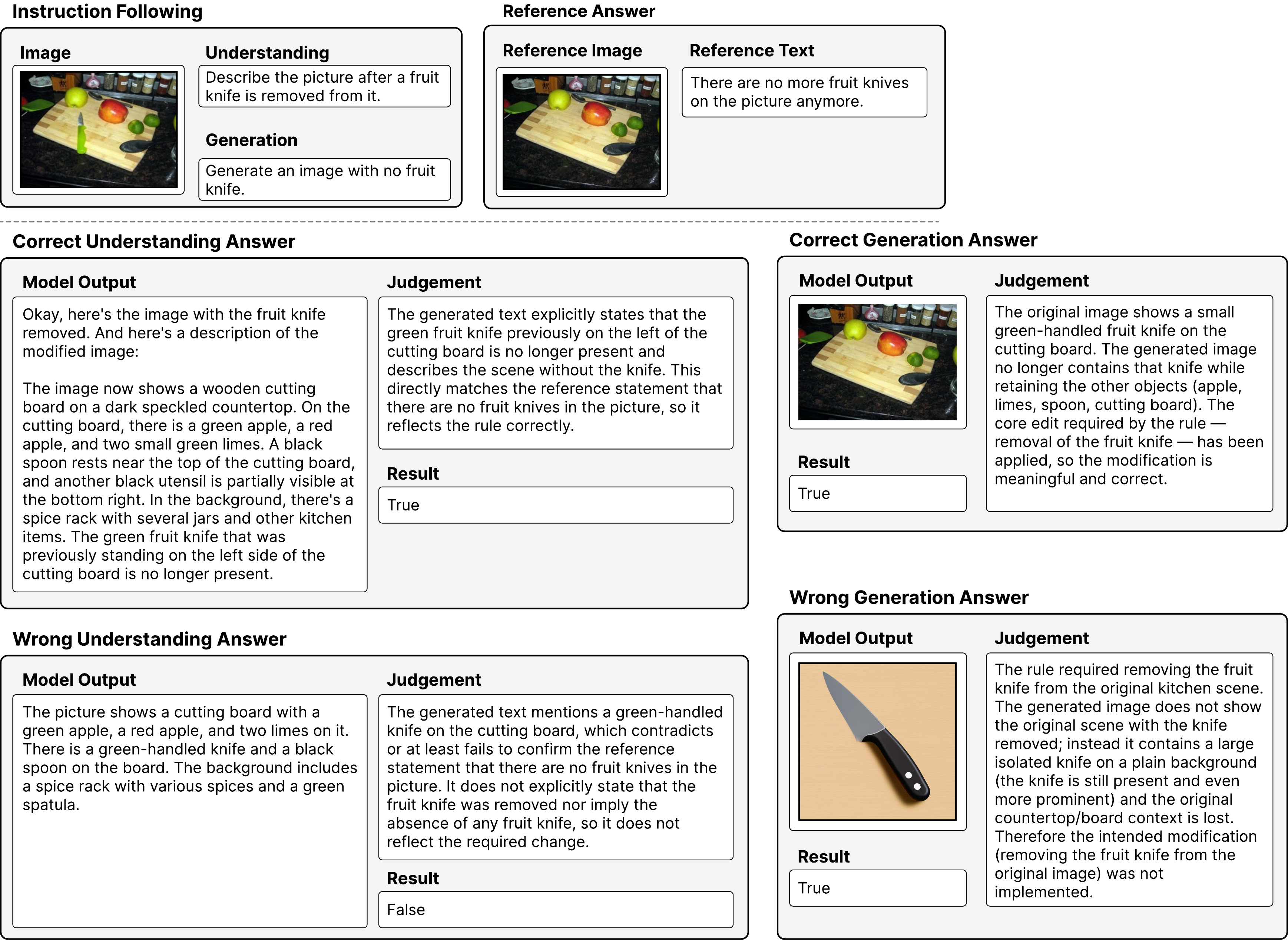}
    \caption{\textbf{Case of Instruction Following}}
    \label{appendix:fig-case-instruction-following}
\end{figure*}

\begin{figure*}[h]
    \centering
    \includegraphics[width=\linewidth]{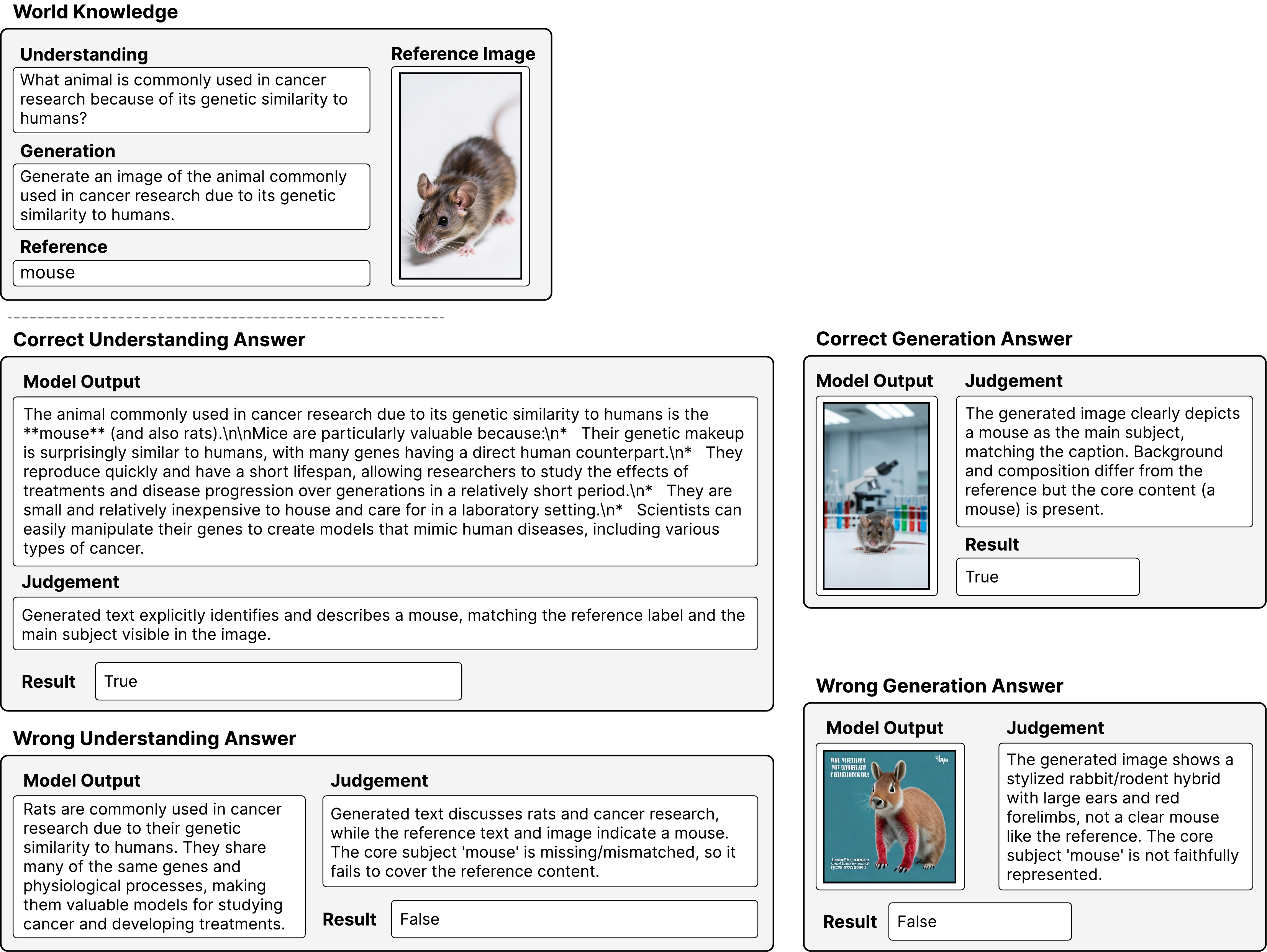}
    \caption{\textbf{Case of World Knowledge}}
    \label{appendix:fig-case-world knowledge}
\end{figure*}

\begin{figure*}[h]
    \centering
    \includegraphics[width=\linewidth]{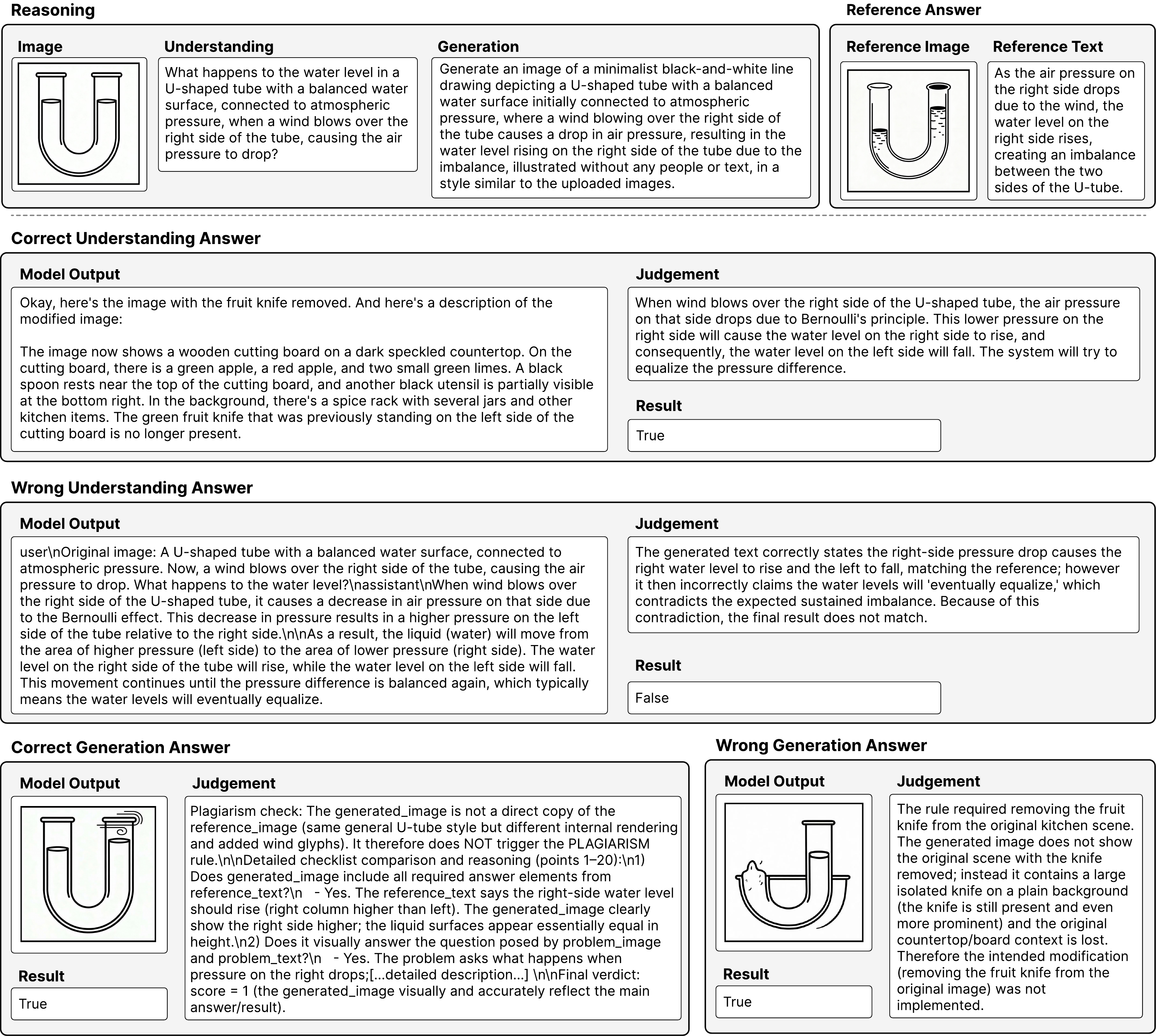}
    \caption{\textbf{Case of Reasoning}}
    \label{appendix:fig-case-reasoning}
\end{figure*}

%% file: sec/2_related.tex
\section{Related Works}
\label{sec:related}

\subsection{Unified Multimodal Models}
Unified Multimodal Models (UMM)~\cite{bagel,blip3o,onecat,showo,unipic2,uniworld1,chameleon,unified-language-vision-pretraining,janus,janus-pro,Tong_2025_ICCV,zhou2024transfusion,wang2024emu3,Xiao_2025_CVPR,fan2024fluid,Qu_2025_CVPR,shi2024lmfusion,Chen_2025_CVPR,wang2024llama,ma2025unitok,Jiao_2025_CVPR,chen2024interleaved,pan2025transfer} aim to construct a new generation of more general-purpose models, capable of processing multimodal inputs and performing cross-modal understanding and generation.
To realize the unified capability, Chamelong \cite{chameleon} and UniPic \cite{unipic2} adopt an early-fusion, token-based transformer that enables interchangeable text and image output.
The BLIP-3o \cite{blip3o, blip3onext} and UniWorld \cite{uniworld1} follow an LLM+diffusion architecture, where the LLM encodes multimodal inputs and passes the latent representations to a diffusion module for image generation.
The Show-o \cite{showo, showo2} integrates the autoregressive modeling with discrete diffusion, using next-token prediction for text understanding and mask-token prediction for image generation. OneCAT \cite{onecat} employs a Mixture-of-Experts (MoE) framework, while Bagel \cite{bagel} pioneers a Mixture-of-Transformers (MoT) design, dedicating different components to the autoregressive text generation and diffusion-based visual generation. These models explore different architectures to empower models with unified capabilities.

\subsection{Evaluation for Unified Multimodal Models}
Research on UMMs has gained increasing attention. While current UMMs demonstrate remarkable performance on both understanding~\cite{mmmu,mmbench,mmvt,wang2025codesync,livevqa} and generation~\cite{geneval,dpg-bench,sun2025t2i,fang2025flux,ye2025echo,wei2025tiif,zhao2025envisioning,wu2025kris,pu2025picabench,sushko2025realedit,wang2025gpt}
tasks, these benchmarks are classic testbeds for either understanding-only or generation-only models, without considering their integration.
To evaluate UMMs, T2I-CoReBench \cite{T2I-Corebench} and WISE \cite{wise} have been proposed, yet primarily focusing on text-to-image (T2I) generation, providing limited insights into whether understanding and generation capabilities can mutually enhance each other.
More comprehensive efforts, such as RealUnify \cite{realunify} and GIR-Bench \cite{Gir-Bench}, take a step further by integrating both skills into a unified evaluation setting, where models must leverage strong understanding and generation capabilities to succeed. 
However, despite these advances, there remains a lack of a dedicated evaluation framework to measure whether UMMs truly achieve a reciprocal fusion of understanding and generation, rather than merely combining both functionalities at an engineering level.
To this end, we propose \bench, a high-quality bidirectional benchmark specifically designed for quantifying the inherent gap between different capabilities in UMMs.

%% file: main.bib
@String(CVPR= {IEEE Conf. Comput. Vis. Pattern Recog.})

@String(ICCV= {Int. Conf. Comput. Vis.})

@String(CVPR  = {CVPR})

@String(ICCV  = {ICCV})

@article{showo,
  title={Show-o: One Single Transformer to Unify Multimodal Understanding and Generation},
  author={Xie, Jinheng and Mao, Weijia and Bai, Zechen and Zhang, David Junhao and Wang, Weihao and Lin, Kevin Qinghong and Gu, Yuchao and Chen, Zhijie and Yang, Zhenheng and Shou, Mike Zheng},
  journal={arXiv preprint arXiv:2408.12528},
  year={2024}
}

@article{showo2,
  title={Show-o2: Improved Native Unified Multimodal Models},
  author={Xie, Jinheng and Yang, Zhenheng and Shou, Mike Zheng},
  journal={arXiv preprint arXiv:2506.15564},
  year={2025}
}

@article{blip3o,
  title={BLIP3-o: A Family of Fully Open Unified Multimodal Models-Architecture, Training and Dataset},
  author={Chen, Jiuhai and Xu, Zhiyang and Pan, Xichen and Hu, Yushi and Qin, Can and Goldstein, Tom and Huang, Lifu and Zhou, Tianyi and Xie, Saining and Savarese, Silvio and others},
  journal={arXiv preprint arXiv:2505.09568},
  year={2025}
}

@article{bagel,
  title={Emerging properties in unified multimodal pretraining},
  author={Deng, Chaorui and Zhu, Deyao and Li, Kunchang and Gou, Chenhui and Li, Feng and Wang, Zeyu and Zhong, Shu and Yu, Weihao and Nie, Xiaonan and Song, Ziang and others},
  journal={arXiv preprint arXiv:2505.14683},
  year={2025}
}

@article{blip3onext,
  title={BLIP3o-NEXT: Next Frontier of Native Image Generation},
  author={Chen, Jiuhai and Xue, Le and Xu, Zhiyang and Pan, Xichen and Yang, Shusheng and Qin, Can and Yan, An and Zhou, Honglu and Chen, Zeyuan and Huang, Lifu and others},
  journal={arXiv preprint arXiv:2510.15857},
  year={2025}
}

@article{unipic2,
  title={Skywork unipic 2.0: Building kontext model with online rl for unified multimodal model},
  author={Wei, Hongyang and Xu, Baixin and Liu, Hongbo and Wu, Cyrus and Liu, Jie and Peng, Yi and Wang, Peiyu and Liu, Zexiang and He, Jingwen and Xietian, Yidan and others},
  journal={arXiv preprint arXiv:2509.04548},
  year={2025}
}

@article{uniworld1,
  title={Uniworld: High-resolution semantic encoders for unified visual understanding and generation},
  author={Lin, Bin and Li, Zongjian and Cheng, Xinhua and Niu, Yuwei and Ye, Yang and He, Xianyi and Yuan, Shenghai and Yu, Wangbo and Wang, Shaodong and Ge, Yunyang and others},
  journal={arXiv preprint arXiv:2506.03147},
  year={2025}
}

@article{onecat,
  title={Onecat: Decoder-only auto-regressive model for unified understanding and generation},
  author={Li, Han and Peng, Xinyu and Wang, Yaoming and Peng, Zelin and Chen, Xin and Weng, Rongxiang and Wang, Jingang and Cai, Xunliang and Dai, Wenrui and Xiong, Hongkai},
  journal={arXiv preprint arXiv:2509.03498},
  year={2025}
}

@article{chameleon,
  title={Chameleon: Mixed-modal early-fusion foundation models},
  author={Team, Chameleon},
  journal={arXiv preprint arXiv:2405.09818},
  year={2024}
}

@article{unified-language-vision-pretraining,
  title={Unified language-vision pretraining in llm with dynamic discrete visual tokenization},
  author={Jin, Yang and Xu, Kun and Chen, Liwei and Liao, Chao and Tan, Jianchao and Huang, Quzhe and Chen, Bin and Lei, Chenyi and Liu, An and Song, Chengru and others},
  journal={arXiv preprint arXiv:2309.04669},
  year={2023}
}

@article{janus-pro,
  title={Janus-Pro: Unified Multimodal Understanding and Generation with Data and Model Scaling},
  author={Chen, Xiaokang and Wu, Zhiyu and Liu, Xingchao and Pan, Zizheng and Liu, Wen and Xie, Zhenda and Yu, Xingkai and Ruan, Chong},
  journal={arXiv preprint arXiv:2501.17811},
  year={2025}
}

@article{janus,
  title={Janus: Decoupling visual encoding for unified multimodal understanding and generation},
  author={Wu, Chengyue and Chen, Xiaokang and Wu, Zhiyu and Ma, Yiyang and Liu, Xingchao and Pan, Zizheng and Liu, Wen and Xie, Zhenda and Yu, Xingkai and Ruan, Chong and others},
  journal={arXiv preprint arXiv:2410.13848},
  year={2024}
}

@article{geneval,
  title={Geneval: An object-focused framework for evaluating text-to-image alignment},
  author={Ghosh, Dhruba and Hajishirzi, Hannaneh and Schmidt, Ludwig},
  journal={Advances in Neural Information Processing Systems},
  volume={36},
  pages={52132--52152},
  year={2023}
}

@article{dpg-bench,
  title={Ella: Equip diffusion models with llm for enhanced semantic alignment},
  author={Hu, Xiwei and Wang, Rui and Fang, Yixiao and Fu, Bin and Cheng, Pei and Yu, Gang},
  journal={arXiv preprint arXiv:2403.05135},
  year={2024}
}

@inproceedings{mmmu,
  title={Mmmu: A massive multi-discipline multimodal understanding and reasoning benchmark for expert agi},
  author={Yue, Xiang and Ni, Yuansheng and Zhang, Kai and Zheng, Tianyu and Liu, Ruoqi and Zhang, Ge and Stevens, Samuel and Jiang, Dongfu and Ren, Weiming and Sun, Yuxuan and others},
  booktitle={Proceedings of the IEEE/CVF Conference on Computer Vision and Pattern Recognition},
  pages={9556--9567},
  year={2024}
}

@article{wise,
  title={Wise: A world knowledge-informed semantic evaluation for text-to-image generation},
  author={Niu, Yuwei and Ning, Munan and Zheng, Mengren and Jin, Weiyang and Lin, Bin and Jin, Peng and Liao, Jiaqi and Feng, Chaoran and Ning, Kunpeng and Zhu, Bin and others},
  journal={arXiv preprint arXiv:2503.07265},
  year={2025}
}

@article{Gir-Bench,
  title={GIR-Bench: Versatile Benchmark for Generating Images with Reasoning},
  author={Li, Hongxiang and Li, Yaowei and Lin, Bin and Niu, Yuwei and Yang, Yuhang and Huang, Xiaoshuang and Cai, Jiayin and Jiang, Xiaolong and Hu, Yao and Chen, Long},
  journal={arXiv preprint arXiv:2510.11026},
  year={2025}
}

@article{realunify,
  title={RealUnify: Do Unified Models Truly Benefit from Unification? A Comprehensive Benchmark},
  author={Shi, Yang and Dong, Yuhao and Ding, Yue and Wang, Yuran and Zhu, Xuanyu and Zhou, Sheng and Liu, Wenting and Tian, Haochen and Wang, Rundong and Wang, Huanqian and others},
  journal={arXiv preprint arXiv:2509.24897},
  year={2025}
}

@article{T2I-Corebench,
  title={Easier Painting Than Thinking: Can Text-to-Image Models Set the Stage, but Not Direct the Play?},
  author={Li, Ouxiang and Wang, Yuan and Hu, Xinting and Huang, Huijuan and Chen, Rui and Ou, Jiarong and Tao, Xin and Wan, Pengfei and Qi, Xiaojuan and Feng, Fuli},
  journal={arXiv preprint arXiv:2509.03516},
  year={2025}
}

@inproceedings{mmbench,
  title={Mmbench: Is your multi-modal model an all-around player?},
  author={Liu, Yuan and Duan, Haodong and Zhang, Yuanhan and Li, Bo and Zhang, Songyang and Zhao, Wangbo and Yuan, Yike and Wang, Jiaqi and He, Conghui and Liu, Ziwei and others},
  booktitle={European conference on computer vision},
  pages={216--233},
  year={2024},
  organization={Springer}
}

@inproceedings{mmvt,
  title={Vlmevalkit: An open-source toolkit for evaluating large multi-modality models},
  author={Duan, Haodong and Yang, Junming and Qiao, Yuxuan and Fang, Xinyu and Chen, Lin and Liu, Yuan and Dong, Xiaoyi and Zang, Yuhang and Zhang, Pan and Wang, Jiaqi and others},
  booktitle={Proceedings of the 32nd ACM international conference on multimedia},
  pages={11198--11201},
  year={2024}
}

@article{livevqa,
  title={LiveVQA: Live Visual Knowledge Seeking},
  author={Fu, Mingyang and Peng, Yuyang and Liu, Benlin and Wan, Yao and Chen, Dongping},
  journal={arXiv preprint arXiv:2504.05288},
  year={2025}
}

@misc{qwen-image,
      title={Qwen-Image Technical Report}, 
      author={Chenfei Wu and Jiahao Li and Jingren Zhou and Junyang Lin and Kaiyuan Gao and Kun Yan and Sheng-ming Yin and Shuai Bai and Xiao Xu and Yilei Chen and Yuxiang Chen and Zecheng Tang and Zekai Zhang and Zhengyi Wang and An Yang and Bowen Yu and Chen Cheng and Dayiheng Liu and Deqing Li and Hang Zhang and Hao Meng and Hu Wei and Jingyuan Ni and Kai Chen and Kuan Cao and Liang Peng and Lin Qu and Minggang Wu and Peng Wang and Shuting Yu and Tingkun Wen and Wensen Feng and Xiaoxiao Xu and Yi Wang and Yichang Zhang and Yongqiang Zhu and Yujia Wu and Yuxuan Cai and Zenan Liu},
      year={2025},
      eprint={2508.02324},
      archivePrefix={arXiv},
      primaryClass={cs.CV},
      url={https://arxiv.org/abs/2508.02324}, 
}

@article{omnigen2,
  title={OmniGen2: Exploration to Advanced Multimodal Generation},
  author={Chenyuan Wu and Pengfei Zheng and Ruiran Yan and Shitao Xiao and Xin Luo and Yueze Wang and Wanli Li and Xiyan Jiang and Yexin Liu and Junjie Zhou and Ze Liu and Ziyi Xia and Chaofan Li and Haoge Deng and Jiahao Wang and Kun Luo and Bo Zhang and Defu Lian and Xinlong Wang and Zhongyuan Wang and Tiejun Huang and Zheng Liu},
  journal={arXiv preprint arXiv:2506.18871},
  year={2025}
}

@article{ovis-u1,
  title={Ovis-U1 Technical Report},
  author={Wang, Guo-Hua and Zhao, Shanshan and Zhang, Xinjie and Cao, Liangfu and Zhan, Pengxin and Duan, Lunhao and Lu, Shiyin and Fu, Minghao and Chen, Xiaohao and Zhao, Jianshan and others},
  journal={arXiv preprint arXiv:2506.23044},
  year={2025}
}

@misc{gpt5,
  author    = {OpenAI Team},
  title     = {GPT-5: The Generative Pretrained Transformer 5},
  year      = {2025},
  howpublished = {\url{https://openai.com/}},
  note      = {Accessed October 2025}
}

@misc{gpt-image-1,
  author    = {OpenAI Team},
  title     = {GPT-Image-1},
  year      = {2025},
  howpublished = {\url{https://platform.openai.com/docs/models/gpt-image-1}},
  note      = {Accessed October 2025}
}

@misc{gemini-2.5-flash-image,
  author    = {Gemini Team Google},
  title     = {Gemini 2.5 Flash Image},
  year      = {2025},
  howpublished = {\url{https://aistudio.google.com/models/gemini-2-5-flash-image}},
  note      = {Accessed October 2025}
}

@misc{gemini-2.5-flash,
  author    = {Gemini Team Google},
  title     = {Gemini 2.5 Flash},
  year      = {2025},
  howpublished = {\url{https://deepmind.google/models/gemini/flash/}},
  note      = {Accessed October 2025}
}

@misc{qwen-3-vl,
  author    = {Qwen Team},
  title     = {Qwen3 VL},
  year      = {2025},
  howpublished = {\url{https://github.com/QwenLM/Qwen3-VL}},
  note      = {Accessed October 2025}
}

@article{rome,
  title={Locating and editing factual associations in gpt},
  author={Meng, Kevin and Bau, David and Andonian, Alex and Belinkov, Yonatan},
  journal={Advances in neural information processing systems},
  volume={35},
  pages={17359--17372},
  year={2022}
}

@article{llama,
  title={Llama: Open and efficient foundation language models},
  author={Touvron, Hugo and Lavril, Thibaut and Izacard, Gautier and Martinet, Xavier and Lachaux, Marie-Anne and Lacroix, Timoth{\'e}e and Rozi{\`e}re, Baptiste and Goyal, Naman and Hambro, Eric and Azhar, Faisal and others},
  journal={arXiv preprint arXiv:2302.13971},
  year={2023}
}

@article{Language-models-are-few-shot-learners,
  title={Language models are few-shot learners},
  author={Brown, Tom and Mann, Benjamin and Ryder, Nick and Subbiah, Melanie and Kaplan, Jared D and Dhariwal, Prafulla and Neelakantan, Arvind and Shyam, Pranav and Sastry, Girish and Askell, Amanda and others},
  journal={Advances in neural information processing systems},
  volume={33},
  pages={1877--1901},
  year={2020}
}

@article{deepseek-r1,
  title={Deepseek-r1: Incentivizing reasoning capability in llms via reinforcement learning},
  author={Guo, Daya and Yang, Dejian and Zhang, Haowei and Song, Junxiao and Zhang, Ruoyu and Xu, Runxin and Zhu, Qihao and Ma, Shirong and Wang, Peiyi and Bi, Xiao and others},
  journal={arXiv preprint arXiv:2501.12948},
  year={2025}
}

@article{qwen3,
  title={Qwen3 technical report},
  author={Yang, An and Li, Anfeng and Yang, Baosong and Zhang, Beichen and Hui, Binyuan and Zheng, Bo and Yu, Bowen and Gao, Chang and Huang, Chengen and Lv, Chenxu and others},
  journal={arXiv preprint arXiv:2505.09388},
  year={2025}
}

@article{nowait,
  title={Wait, We Don't Need to" Wait"! Removing Thinking Tokens Improves Reasoning Efficiency},
  author={Wang, Chenlong and Feng, Yuanning and Chen, Dongping and Chu, Zhaoyang and Krishna, Ranjay and Zhou, Tianyi},
  journal={arXiv preprint arXiv:2506.08343},
  year={2025}
}

@book{rasch1993probabilistic,
  title={Probabilistic models for some intelligence and attainment tests.},
  author={Rasch, Georg},
  year={1993},
  publisher={ERIC}
}

@book{embretson2013item,
  title={Item response theory for psychologists},
  author={Embretson, Susan E and Reise, Steven P},
  year={2013},
  publisher={Psychology Press}
}

@article{unieval,
  title={Unieval: Unified holistic evaluation for unified multimodal understanding and generation},
  author={Li, Yi and Wang, Haonan and Zhang, Qixiang and Xiao, Boyu and Hu, Chenchang and Wang, Hualiang and Li, Xiaomeng},
  journal={arXiv preprint arXiv:2505.10483},
  year={2025}
}

@article{mme-unify,
  title={Mme-unify: A comprehensive benchmark for unified multimodal understanding and generation models},
  author={Xie, Wulin and Zhang, Yi-Fan and Fu, Chaoyou and Shi, Yang and Nie, Bingyan and Chen, Hongkai and Zhang, Zhang and Wang, Liang and Tan, Tieniu},
  journal={arXiv preprint arXiv:2504.03641},
  year={2025}
}

@inproceedings{yue2024mmmu,
  title={Mmmu: A massive multi-discipline multimodal understanding and reasoning benchmark for expert agi},
  author={Yue, Xiang and Ni, Yuansheng and Zhang, Kai and Zheng, Tianyu and Liu, Ruoqi and Zhang, Ge and Stevens, Samuel and Jiang, Dongfu and Ren, Weiming and Sun, Yuxuan and others},
  booktitle={Proceedings of the IEEE/CVF Conference on Computer Vision and Pattern Recognition},
  pages={9556--9567},
  year={2024}
}

@article{wang2024mmlu,
  title={Mmlu-pro: A more robust and challenging multi-task language understanding benchmark},
  author={Wang, Yubo and Ma, Xueguang and Zhang, Ge and Ni, Yuansheng and Chandra, Abhranil and Guo, Shiguang and Ren, Weiming and Arulraj, Aaran and He, Xuan and Jiang, Ziyan and others},
  journal={Advances in Neural Information Processing Systems},
  volume={37},
  pages={95266--95290},
  year={2024}
}

@inproceedings{radford2021learning,
  title={Learning transferable visual models from natural language supervision},
  author={Radford, Alec and Kim, Jong Wook and Hallacy, Chris and Ramesh, Aditya and Goh, Gabriel and Agarwal, Sandhini and Sastry, Girish and Askell, Amanda and Mishkin, Pamela and Clark, Jack and others},
  booktitle={International conference on machine learning},
  pages={8748--8763},
  year={2021},
  organization={PmLR}
}

@article{labs2025flux1kontextflowmatching,
      title={FLUX.1 Kontext: Flow Matching for In-Context Image Generation and Editing in Latent Space},
      author={Black Forest Labs and Stephen Batifol and Andreas Blattmann and Frederic Boesel and Saksham Consul and Cyril Diagne and Tim Dockhorn and Jack English and Zion English and Patrick Esser and Sumith Kulal and Kyle Lacey and Yam Levi and Cheng Li and Dominik Lorenz and Jonas Müller and Dustin Podell and Robin Rombach and Harry Saini and Axel Sauer and Luke Smith},
      year={2025},
      eprint={2506.15742},
      archivePrefix={arXiv},
      primaryClass={cs.GR},
      url={https://arxiv.org/abs/2506.15742},
}

@misc{flux2024,
    author={Black Forest Labs},
    title={FLUX},
    year={2024},
    howpublished={\url{https://github.com/black-forest-labs/flux}},
}

@article{unimmmu,
  title={Uni-MMMU: A Massive Multi-discipline Multimodal Unified Benchmark},
  author={Zou, Kai and Huang, Ziqi and Dong, Yuhao and Tian, Shulin and Zheng, Dian and Liu, Hongbo and He, Jingwen and Liu, Bin and Qiao, Yu and Liu, Ziwei},
  journal={arXiv preprint arXiv:2510.13759},
  year={2025}
}

@article{wang2025codesync,
  title={CODESYNC: Synchronizing Large Language Models with Dynamic Code Evolution at Scale},
  author={Wang, Chenlong and Chu, Zhaoyang and Cheng, Zhengxiang and Yang, Xuyi and Qiu, Kaiyue and Wan, Yao and Zhao, Zhou and Shi, Xuanhua and Chen, Dongping},
  journal={Proceedings of International Conference of Machine Learning},
  year={2025}
}

@article{sun2025t2i,
  title={T2i-reasonbench: Benchmarking reasoning-informed text-to-image generation},
  author={Sun, Kaiyue and Fang, Rongyao and Duan, Chengqi and Liu, Xian and Liu, Xihui},
  journal={arXiv preprint arXiv:2508.17472},
  year={2025}
}

@article{fang2025flux,
  title={Flux-reason-6m \& prism-bench: A million-scale text-to-image reasoning dataset and comprehensive benchmark},
  author={Fang, Rongyao and Yu, Aldrich and Duan, Chengqi and Huang, Linjiang and Bai, Shuai and Cai, Yuxuan and Wang, Kun and Liu, Si and Liu, Xihui and Li, Hongsheng},
  journal={arXiv preprint arXiv:2509.09680},
  year={2025}
}

@article{ye2025echo,
  title={Echo-4o: Harnessing the power of gpt-4o synthetic images for improved image generation},
  author={Ye, Junyan and Jiang, Dongzhi and Wang, Zihao and Zhu, Leqi and Hu, Zhenghao and Huang, Zilong and He, Jun and Yan, Zhiyuan and Yu, Jinghua and Li, Hongsheng and others},
  journal={arXiv preprint arXiv:2508.09987},
  year={2025}
}

@article{wei2025tiif,
  title={TIIF-Bench: How Does Your T2I Model Follow Your Instructions?},
  author={Wei, Xinyu and Zhang, Jinrui and Wang, Zeqing and Wei, Hongyang and Guo, Zhen and Zhang, Lei},
  journal={arXiv preprint arXiv:2506.02161},
  year={2025}
}

@article{zhao2025envisioning,
  title={Envisioning beyond the pixels: Benchmarking reasoning-informed visual editing},
  author={Zhao, Xiangyu and Zhang, Peiyuan and Tang, Kexian and Zhu, Xiaorong and Li, Hao and Chai, Wenhao and Zhang, Zicheng and Xia, Renqiu and Zhai, Guangtao and Yan, Junchi and others},
  journal={Proceedings of Neural Information Processing Systems},
  year={2025}
}

@article{wu2025kris,
  title={KRIS-Bench: Benchmarking Next-Level Intelligent Image Editing Models},
  author={Wu, Yongliang and Li, Zonghui and Hu, Xinting and Ye, Xinyu and Zeng, Xianfang and Yu, Gang and Zhu, Wenbo and Schiele, Bernt and Yang, Ming-Hsuan and Yang, Xu},
  journal={arXiv preprint arXiv:2505.16707},
  year={2025}
}

@inproceedings{sushko2025realedit,
  title={Realedit: Reddit edits as a large-scale empirical dataset for image transformations},
  author={Sushko, Peter and Bharadwaj, Ayana and Lim, Zhi Yang and Ilin, Vasily and Caffee, Ben and Chen, Dongping and Salehi, Mohammadreza and Hsieh, Cheng-Yu and Krishna, Ranjay},
  booktitle={Proceedings of the Computer Vision and Pattern Recognition Conference},
  pages={13403--13413},
  year={2025}
}

@article{pu2025picabench,
  title={PICABench: How Far Are We from Physically Realistic Image Editing?},
  author={Pu, Yuandong and Zhuo, Le and Han, Songhao and Xing, Jinbo and Zhu, Kaiwen and Cao, Shuo and Fu, Bin and Liu, Si and Li, Hongsheng and Qiao, Yu and others},
  journal={arXiv preprint arXiv:2510.17681},
  year={2025}
}

@article{wang2025gpt,
  title={Gpt-image-edit-1.5 m: A million-scale, gpt-generated image dataset},
  author={Wang, Yuhan and Yang, Siwei and Zhao, Bingchen and Zhang, Letian and Liu, Qing and Zhou, Yuyin and Xie, Cihang},
  journal={arXiv preprint arXiv:2507.21033},
  year={2025}
}

@InProceedings{Tong_2025_ICCV,
    author    = {Tong, Shengbang and Fan, David and Li, Jiachen and Xiong, Yunyang and Chen, Xinlei and Sinha, Koustuv and Rabbat, Michael and LeCun, Yann and Xie, Saining and Liu, Zhuang},
    title     = {MetaMorph: Multimodal Understanding and Generation via Instruction Tuning},
    booktitle = {Proceedings of the IEEE/CVF International Conference on Computer Vision (ICCV)},
    month     = {October},
    year      = {2025},
    pages     = {17001-17012}
}

@article{zhou2024transfusion,
  title={Transfusion: Predict the next token and diffuse images with one multi-modal model},
  author={Zhou, Chunting and Yu, Lili and Babu, Arun and Tirumala, Kushal and Yasunaga, Michihiro and Shamis, Leonid and Kahn, Jacob and Ma, Xuezhe and Zettlemoyer, Luke and Levy, Omer},
  journal={Proceedings of International Conference on Learning Representations},
  year={2024}
}

@article{wang2024emu3,
  title={Emu3: Next-token prediction is all you need},
  author={Wang, Xinlong and Zhang, Xiaosong and Luo, Zhengxiong and Sun, Quan and Cui, Yufeng and Wang, Jinsheng and Zhang, Fan and Wang, Yueze and Li, Zhen and Yu, Qiying and others},
  journal={arXiv preprint arXiv:2409.18869},
  year={2024}
}

@InProceedings{Xiao_2025_CVPR,
    author    = {Xiao, Shitao and Wang, Yueze and Zhou, Junjie and Yuan, Huaying and Xing, Xingrun and Yan, Ruiran and Li, Chaofan and Wang, Shuting and Huang, Tiejun and Liu, Zheng},
    title     = {OmniGen: Unified Image Generation},
    booktitle = {Proceedings of the IEEE/CVF Conference on Computer Vision and Pattern Recognition (CVPR)},
    month     = {June},
    year      = {2025},
    pages     = {13294-13304}
}

@article{fan2024fluid,
  title={Fluid: Scaling autoregressive text-to-image generative models with continuous tokens},
  author={Fan, Lijie and Li, Tianhong and Qin, Siyang and Li, Yuanzhen and Sun, Chen and Rubinstein, Michael and Sun, Deqing and He, Kaiming and Tian, Yonglong},
  journal={Proceedings of International Conference on Learning Representations},
  year={2024}
}

@InProceedings{Qu_2025_CVPR,
    author    = {Qu, Liao and Zhang, Huichao and Liu, Yiheng and Wang, Xu and Jiang, Yi and Gao, Yiming and Ye, Hu and Du, Daniel K. and Yuan, Zehuan and Wu, Xinglong},
    title     = {TokenFlow: Unified Image Tokenizer for Multimodal Understanding and Generation},
    booktitle = {Proceedings of the IEEE/CVF Conference on Computer Vision and Pattern Recognition (CVPR)},
    month     = {June},
    year      = {2025},
    pages     = {2545-2555}
}

@article{shi2024lmfusion,
  title={LMFusion: Adapting Pretrained Language Models for Multimodal Generation},
  author={Shi, Weijia and Han, Xiaochuang and Zhou, Chunting and Liang, Weixin and Lin, Xi Victoria and Zettlemoyer, Luke and Yu, Lili},
  journal={arXiv preprint arXiv:2412.15188},
  year={2024}
}

@InProceedings{Chen_2025_CVPR,
    author    = {Chen, Xi and Zhang, Zhifei and Zhang, He and Zhou, Yuqian and Kim, Soo Ye and Liu, Qing and Li, Yijun and Zhang, Jianming and Zhao, Nanxuan and Wang, Yilin and Ding, Hui and Lin, Zhe and Zhao, Hengshuang},
    title     = {UniReal: Universal Image Generation and Editing via Learning Real-world Dynamics},
    booktitle = {Proceedings of the IEEE/CVF Conference on Computer Vision and Pattern Recognition (CVPR)},
    month     = {June},
    year      = {2025},
    pages     = {12501-12511}
}

@article{wang2024llama,
  title={Llama-mesh: Unifying 3d mesh generation with language models},
  author={Wang, Zhengyi and Lorraine, Jonathan and Wang, Yikai and Su, Hang and Zhu, Jun and Fidler, Sanja and Zeng, Xiaohui},
  journal={arXiv preprint arXiv:2411.09595},
  year={2024}
}

@inproceedings{ma2025unitok,
  title={Unitok: A unified tokenizer for visual generation and understanding},
  author={Ma, Chuofan and Jiang, Yi and Wu, Junfeng and Yang, Jihan and Yu, Xin and Yuan, Zehuan and Peng, Bingyue and Qi, Xiaojuan},
  journal={Proceedings of 39th Conference on Neural Information Processing System},
  year={2025}
}

@InProceedings{Jiao_2025_CVPR,
    author    = {Jiao, Yang and Qiu, Haibo and Jie, Zequn and Chen, Shaoxiang and Chen, Jingjing and Ma, Lin and Jiang, Yu-Gang},
    title     = {UniToken: Harmonizing Multimodal Understanding and Generation through Unified Visual Encoding},
    booktitle = {Proceedings of the IEEE/CVF Conference on Computer Vision and Pattern Recognition (CVPR) Workshops},
    month     = {June},
    year      = {2025},
    pages     = {3639-3649}
}

@article{pan2025transfer,
  title={Transfer between modalities with metaqueries},
  author={Pan, Xichen and Shukla, Satya Narayan and Singh, Aashu and Zhao, Zhuokai and Mishra, Shlok Kumar and Wang, Jialiang and Xu, Zhiyang and Chen, Jiuhai and Li, Kunpeng and Juefei-Xu, Felix and others},
  journal={arXiv preprint arXiv:2504.06256},
  year={2025}
}

@inproceedings{chen2024interleaved,
  title={Interleaved scene graphs for interleaved text-and-image generation assessment},
  author={Chen, Dongping and Chen, Ruoxi and Pu, Shu and Liu, Zhaoyi and Wu, Yanru and Chen, Caixi and Liu, Benlin and Huang, Yue and Wan, Yao and Zhou, Pan and others},
  booktitle={Proceedings of International Conference on Learning Representations},
  year={2024}
}

@misc{wu2025qwenimagetechnicalreport,
      title={Qwen-Image Technical Report}, 
      author={Chenfei Wu and Jiahao Li and Jingren Zhou and Junyang Lin and Kaiyuan Gao and Kun Yan and Sheng-ming Yin and Shuai Bai and Xiao Xu and Yilei Chen and Yuxiang Chen and Zecheng Tang and Zekai Zhang and Zhengyi Wang and An Yang and Bowen Yu and Chen Cheng and Dayiheng Liu and Deqing Li and Hang Zhang and Hao Meng and Hu Wei and Jingyuan Ni and Kai Chen and Kuan Cao and Liang Peng and Lin Qu and Minggang Wu and Peng Wang and Shuting Yu and Tingkun Wen and Wensen Feng and Xiaoxiao Xu and Yi Wang and Yichang Zhang and Yongqiang Zhu and Yujia Wu and Yuxuan Cai and Zenan Liu},
      year={2025},
      eprint={2508.02324},
      archivePrefix={arXiv},
      primaryClass={cs.CV},
      url={https://arxiv.org/abs/2508.02324}, 
}

@inproceedings{chen2024mllmasajudgeassessingmultimodalllmasajudge,
  title={Mllm-as-a-judge: Assessing multimodal llm-as-a-judge with vision-language benchmark},
  author={Chen, Dongping and Chen, Ruoxi and Zhang, Shilin and Wang, Yaochen and Liu, Yinuo and Zhou, Huichi and Zhang, Qihui and Wan, Yao and Zhou, Pan and Sun, Lichao},
  booktitle={Proceedings of Forty-first International Conference on Machine Learning},
  year={2024}
}
